# Drake: An Efficient Executive for Temporal Plans with Choice

**Patrick R. Conrad**                                    PRCONRAD@MIT.EDU
**Brian C. Williams**                                    WILLIAMS@MIT.EDU
*Room 32-227*
*32 Vassar St*
*Cambridge, MA 02139 USA*

## Abstract

This work presents Drake, a dynamic executive for temporal plans with choice. Dynamic plan execution strategies allow an autonomous agent to react quickly to unfolding events, improving the robustness of the agent. Prior work developed methods for dynamically dispatching Simple Temporal Networks, and further research enriched the expressiveness of the plans executives could handle, including discrete choices, which are the focus of this work. However, in some approaches to date, these additional choices induce significant storage or latency requirements to make flexible execution possible.

Drake is designed to leverage the low latency made possible by a preprocessing step called *compilation*, while avoiding high memory costs through a compact representation. We leverage the concepts of labels and environments, taken from prior work in Assumption-based Truth Maintenance Systems (ATMS), to concisely record the implications of the discrete choices, exploiting the structure of the plan to avoid redundant reasoning or storage. Our labeling and maintenance scheme, called the Labeled Value Set Maintenance System, is distinguished by its focus on properties fundamental to temporal problems, and, more generally, weighted graph algorithms. In particular, the maintenance system focuses on maintaining a minimal representation of *non-dominated* constraints. We benchmark Drake's performance on random structured problems, and find that Drake reduces the size of the compiled representation by a factor of over 500 for large problems, while incurring only a modest increase in run-time latency, compared to prior work in compiled executives for temporal plans with discrete choices.

## 1. Introduction

Model-based executives elevate commanding of autonomous systems to the level of goal states while providing guarantees of correctness (Williams, Ingham, Chung, & Elliott, 2003). Using a model-based executive, a user can provide a specification of the goal behavior of the robot and leave it to a program, the *executive*, to determine an appropriate course of action that meets those goals. Temporal plan executives are designed to work with plans including timing requirements.

Typically, for an executive to be robust to disturbances, it must be able to react to the outcomes of events on the fly, otherwise, even seemingly inconsequential variations in the outcomes of events may cause a failure. Thus, it can be helpful to follow a strategy of least commitment and delay each decision until it is actually time to act on that decision, allowing the executive to act with as much information as possible. In the case of temporal plans, an executive following this strategy is said to *dynamically dispatch* the plan (Muscettola,

Morris, & Tsamardinos, 1998). Such an executive is responsible for determining when to schedule events as late as possible while guaranteeing that a consistent schedule exists for all the remaining events. If an external disturbance causes some timing requirement to be violated, then the executive should discover the failure and signal it as soon as possible.

Making such decisions on the fly requires some care, as on-line temporal reasoning can introduce latency that is unacceptable for a real-time system. Therefore, Muscettola et al. (1998) developed a low-latency executive for Simple Temporal Networks (STNs), where an STN is comprised of a set of events and difference constraints on the time of execution of the events. To achieve low latency, the executive is broken into two parts, a *compiler* and a *dispatcher*. The compiler is run in advance to discover and explicitly record all temporal constraints that cannot be quickly inferred on-line, thereby computing the *dispatchable form* of the plan. The dispatcher uses this form to make real-time decisions using a greedy strategy and local, low-latency inferences.

While dynamic scheduling has proven effective, robustness can further be improved by making additional decision dynamically, such as the assignment of an activity to a particular resource. Encoding these decisions requires a more expressive formalism than STNs. Consequently, subsequent research has developed efficient executives for more expressive frameworks, many of which are variants of the STN. Examples of added features include explicit modeling of uncertainty (Morris, Muscettola, & Vidal, 2001; Rossi, Venable, & Yorke-Smith, 2006; Shah & Williams, 2008), discrete choices (Kim, Williams, & Abramson, 2001; Tsamardinos, Pollack, & Ganchev, 2001; Combi & Posenato, 2009; Shah & Williams, 2008), preferences (Hiatt, Zimmerman, Smith, & Simmons, 2009; Khatib, Morris, Morris, & Rossi, 2001; Kim et al., 2001), discrete observations (Tsamardinos, Vidal, & Pollack, 2003), and combinations thereof.

This work focuses on enriching the executive to simultaneously schedule events and make discrete choices as the execution unfolds. The ability to make discrete choices greatly enriches an executive by offering it the ability to dynamically allocate resources, order activities, and choose between alternate methods (sub-plans) for achieving goals. Although prior works have developed executives for this type of plan, they have trade-offs in performance. For example, Tsamardinos, Pollack, and Ganchev (2001) presented an executive for Disjunctive Temporal Networks (DTN), a variant of STNs that include discrete choices. Their executive extends the compilation strategy for STNs by breaking the DTN into its complete exponential set of component STNs and then compiling and dispatching each in parallel. Their strategy offers low latency, but incurs a high storage cost for the dispatchable plan. Another example, Kirk, is an executive for Temporal Plan Networks (TPNs), which extends STNs by including a hierarchical choice between sub-plans, developed by Kim, Williams, and Abramson (2001). Kirk selects a set of choices and performs incremental re-planning whenever a disturbance invalidates that choice, leaving a small memory footprint, but potentially inducing high latency when it selects new choices. Chaski is an executive presented by Shah and Williams (2007) for temporal plans with resource allocation, whose expressiveness is between that of STNs and DTNs. Chaski takes an approach which is a hybrid of the incremental strategy of Kirk and the compiled approach of Tsamardinos et al.: its compiled representation is a base plan and a set of incremental differences, which provides the benefits of compiled execution while improving efficiency by exploiting structure in the plan .

We develop Drake, a novel executive for temporal plans with choice encoded using the expressive representation of DTNs (DTNs dominate TPNs, TCNs, and STNs). Drake achieves low run-time latency through compilation, yet requires less storage than the fully exponential expansion approach taken by Tsamardinos et al. (2001). In order to accomplish this, Drake works on a compact representation of the temporal constraints and discrete choices. To develop the compact representation, we begin with the idea, taken from truth maintenance, of *labeling* consequences of inferences with the minimal set of choices that imply the consequence; this minimal set is called an *environment* (McDermott, 1983; de Kleer, 1986). By monotonicity of inference, this consequence also holds for all sets of choices that are a superset of the minimal environment, thus the environment is a compact encoding of the decision contexts in which that consequence holds.

These ideas are directly applicable to temporal reasoning problems; Drake extends them by leveraging properties fundamental to temporal reasoning problems, and weighted graph problems in general. More specifically, temporal reasoning is non-monotonic, in the sense that it does not need to explicitly represent all derivable constraints, only the tightest possible ones, referred to as the *non-dominated constraints*. Drake uses this property throughout to reduce the computations and storage required. For example, in the inequality $A \leq 4 \leq 8$, where $A$ is a temporal event, there is no need to store the constraint $A \leq 8$, as the tighter inequality makes it unnecessary, or *dominates* it. The focus on non-dominated values or constraints is central to a range of inference problems, including temporal inference, interval reasoning, and inference over weighted graphs.

Dechter, Meiri, and Pearl (1991) proved that STN inference problems are reducible to a widely used set of inference methods on weighted graphs, such as Single Source Shortest Path and All-Pairs Shortest Path Problems. Our approach is to develop labeled analogues to the weighted graph structures that support these shortest path algorithms, providing a compact representation for Drake. In this paper, we first present a new formalism for plans with choice, the Labeled Simple Temporal Network (Labeled STN), which has the same expressiveness as previous formalisms, but which shares ideas with the rest of our techniques. Second, we explain a system for maintaining and deriving compact representations of values that vary with choices, called the Labeled Value Set Maintenance System. Then, we use Labeled Value Sets to construct Labeled Distance Graphs, which are distance graphs where the edge weights may vary depending on the discrete choices. Finally, Drake's compilation and dispatching algorithms are built around these techniques. While the focus of this paper is on dispatchable execution, the techniques surrounding labeled distance graphs hold the promise of extending a wide range of reasoning methods involving graph algorithms to include choice.

In practical terms, Drake's compact encoding provides a reduction in the size of the plan used by the dispatcher by over two orders of magnitude for problems with around 2,000 component STNs, as compared to Tsamardinos et al.'s work (2001). This size reduction comes at a modest increase in the run-time latency, making Drake a useful addition to the available executives.

## 1.1 Overview of the Problem

Drake takes as its input a Labeled STN, which is a temporal constraint representation with choices; in Section 3 we discuss how Labeled STNs can encode choices between sub-plans, temporal constraints, and resource assignment, and mappings to related frameworks. Drake's output is a dynamic execution of the plan, where it determines in real-time when to execute the events, such that at the end of the plan the execution times are consistent with every temporal constraint implied by at least one complete set of choices, barring unforeseen disturbances. If outside disturbances make every possible execution inconsistent, then Drake signals the failure as soon as all possible solutions are rendered inconsistent.

Section 3 provides a formal definition of Labeled STNs; essentially, it is a collection of events to schedule and the constraints the executive must follow. The events may be constrained with simple temporal constraints, which limit the difference in the scheduled times of two events. Furthermore, the Labeled STN specifies discrete choices, where assignments to the choices may imply additional simple temporal constraints.

Throughout this paper, we use the following simple example, which includes a choice between sub-plans.

**Example 1.1** A rover has 100 minutes to work before a scheduled contact with its operators. Before contact, the rover must drive to the next landmark, taking between 30 and 70 minutes. To fill any remaining time, the rover has two options: collect some samples or charge its batteries. Collecting samples consistently takes 50 to 60 minutes, whereas charging the batteries can be usefully done for any duration up to 50 minutes.  □

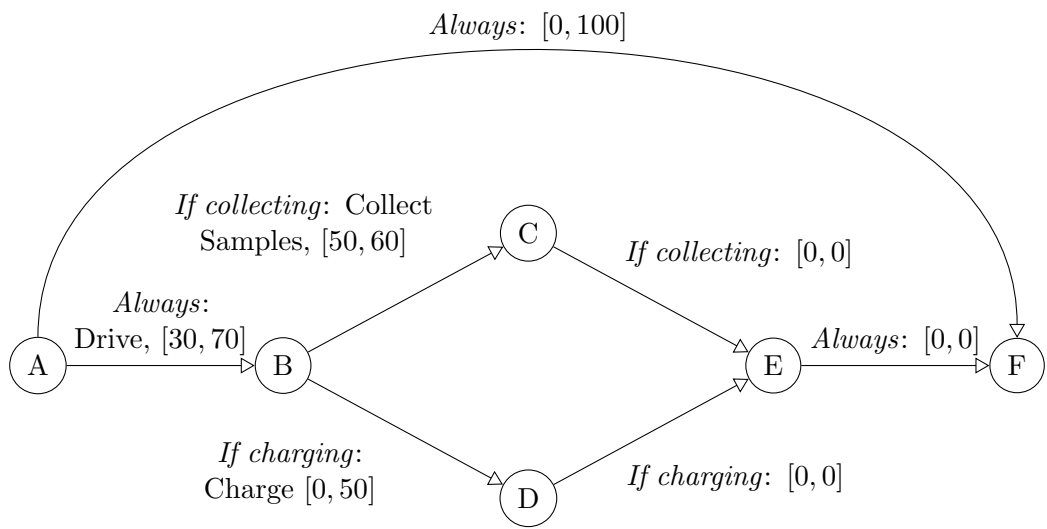

Figure 1.1: This informal Labeled STN depicts Example 1.1. The rover needs to drive, then either collect samples or charge its batteries within a certain time limit.

Figure 1.1 shows an informal representation of the Labeled-STN corresponding to this plan. The notation $[l, u]$ on an edge from vertex $X$ to $Y$ means that the difference in the execution times of the events lies in the given interval, which expresses the constraint $l \leq Y - X \leq u$. In this figure the text explains which temporal constraints are implied by which choice; we develop precise notation later. There are two types of constraints drawn, those that are always required, and those only required if the rover is either collecting samples or charging.

Consider the following correct output execution sequence for the rover problem. In this example, we focus on the form of the executive's output, deferring the presentation of the decision-making strategy until later.

**Example 1.2** Drake starts executing the plan, arbitrarily denoting the starting time as $t = 0$. At that time, it instructs the system to begin the drive activity, indicating that the drive should take 40 minutes. The executive then waits until the system responds that that the drive has completed, at time $t = 45$. Then Drake selects the sample collection option, which had not been determined before, and initiates the activity with a duration of 50 minutes. At $t = 95$, the sample collection completes, finishing the plan within the time limit of 100 minutes. □

## 1.2 Approach: Exploiting Shared Structure through Labeling

Drake's strategy during compilation is to begin with the Labeled STN, a concise statement of the temporal constraints and the choices in the plan. Then, Drake constructs the Labeled Distance Graph associated with the Labeled STN, yielding a single graph structure representing all the possible choices and constraints. Next, Drake's compiler computes the dispatchable form of the problem, which is also a Labeled Distance Graph. This compilation is performed in a unified process that is able to exploit any similarities between the choices to make the representation compact. In contrast, the prior work of Tsamardinos et al. (2001), breaks the input plan into independent STNs, hence their compilation strategy cannot exploit any similarities or shared structures between the choices. There are pathological cases, where every choice is completely unrelated, where there are no similarities for Drake to exploit. However, we expect that nearly every real-world or human designed plan has some degree of shared structure, because a plan usually has some unifying idea which the choices are designed to accomplish. Indeed, we expect that most real plans have significant similarities, allowing Drake to perform well. This section gives an overview of the intuition behind the representation Drake uses and the similarities that Drake exploit.

Before we continue discussing the rover example, consider Figure 1.2, which depicts a small STN, its associated distance graph, and the dispatchable distance graph that is the result of compilation. The set of events in the STN are represented as vertices in the distance graph. Upper bounds in the STN induce edges in the forward direction, weighted with the upper bound, and lower bounds induce edges in the reverse direction weighted with the negative of the lower bound. The distance graph in Figure 1.2b is compiled, and the compiler outputs a new distance graph that contains representations of the constraints needed by the dispatcher. The dispatchable form in Figure 1.2c is used by a dispatcher; execution times are propagated through the edges to determine when other events may be executed.

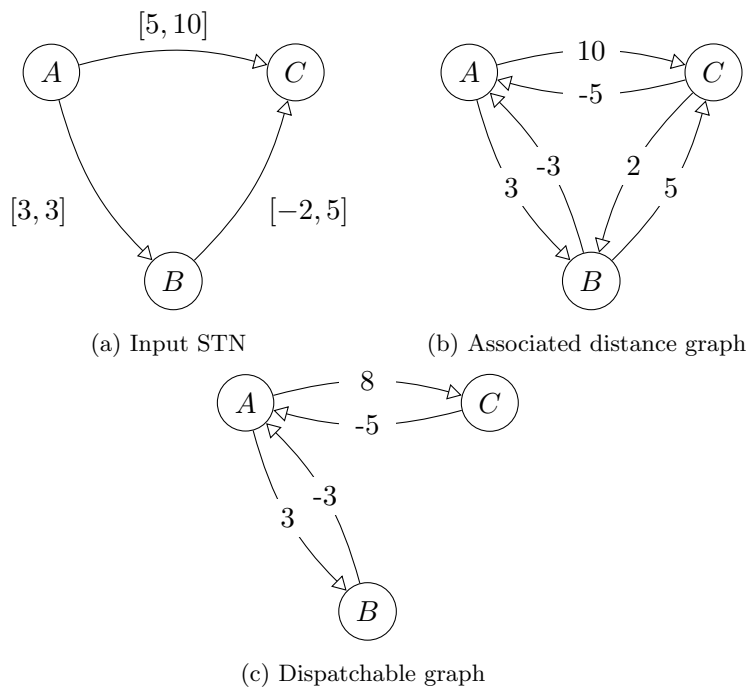

(a) Input STN

(b) Associated distance graph

(c) Dispatchable graph

Figure 1.2: A simple example of reformulating an STN into its associated distance graph and its dispatchable form.

The rover example has a single binary choice, hence for this problem Tsamardinos et al.'s (2001) algorithm separates the two possible STNs then compute their associated distance graphs, which are shown in Figure 1.3. Note the repetition of certain edges in both graphs, for example, the edge $A \to F$, which is present throughout their compilation and dispatch process. Plans with more choices can have an exponential number of repetitions, which can be costly, and which Drake is designed to eliminate.

An informal version of the Labeled Distance Graph associated with the rover example is shown in Figure 1.4. The differing constraints that result from the two possible assignments to the choice are distinguished by annotations called *labels*. For example, edge $(B, C)$ has weight $S : 60$, which indicates whenever sampling is chosen, the edge has a weight of at most 60; the value 60 is labeled with the discrete choice, $S$, that implies it. We gather all the possible values under all the choices into Labeled Value Sets, which are placed on edges. In this example, each edge has a Labeled Value Set with exactly one labeled value, although this is not true in general. A Labeled Distance Graph is essentially a distance graph where the numeric weights are replaced with Labeled Value Sets. Although we develop more precise notation later on in this article, this version shows the intuition behind the approach. Drake capitalizes on this improvement by using the compact representation throughout compilation and dispatch, and this work develops the necessary machinery.

This paper is organized as follows. Section 2 discusses related work on temporal executives and provides background on truth maintenance. Section 3 defines Labeled STNs and their correct dynamic execution, specifying the problem Drake solves. Section 4 recalls the link between STNs and distance graphs, and provides the labeled version of distance graphs, which Drake uses for reasoning. Section 5 presents the Labeled Value Set Maintenance System, completing the foundation of labeled techniques. Section 6 details the dispatcher and Section 7 develops Drake's compilation algorithm. Finally, Section 8 provides some theoretical and experimental performance results, and Section 9 gives some concluding remarks.

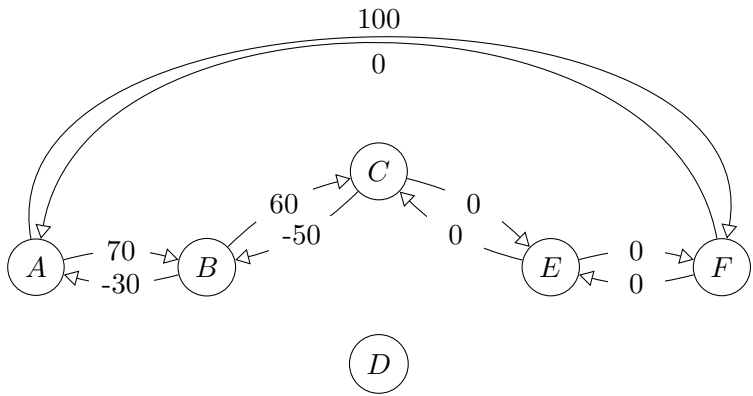

(a) Distance graph if collecting samples

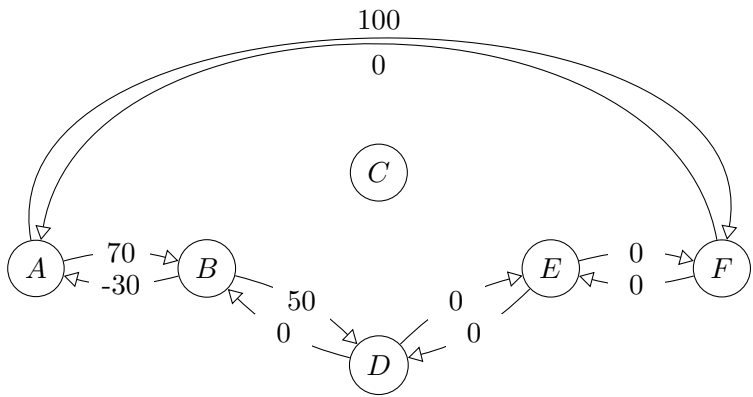

(b) Distance graph if charging

Figure 1.3: The Tsamardinos et al. (2001) style distance graphs associated with the Labeled-STN in Figure 1.1.

## 2. Related Work

Before developing Drake, we give an overview of some relevant literature in the two major areas Drake draws from: scheduling frameworks and truth maintenance.

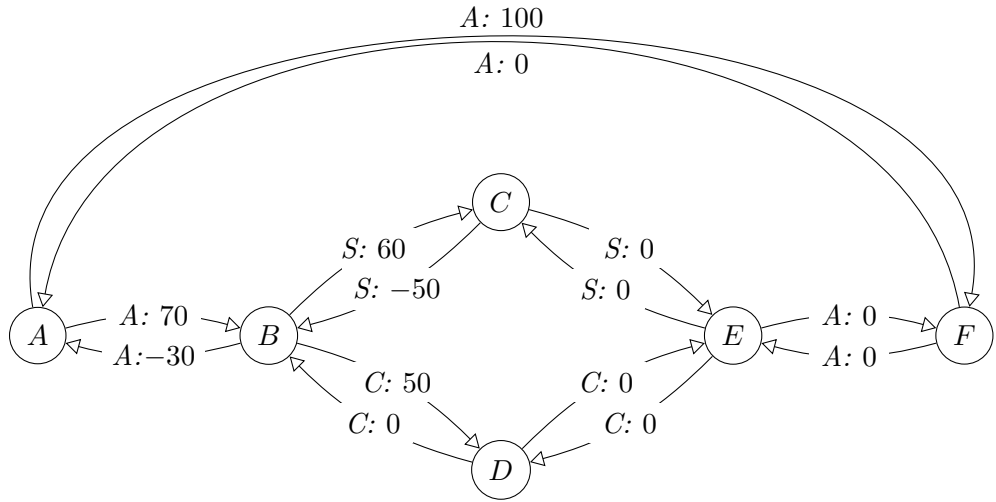

Figure 1.4: An informal labeled distance graph for the rover example. *A:, S:* and *C:,* correspond to weights that hold always, with sampling, and with charging, respectively.

## 2.1 Scheduling Frameworks and Executives

As stated in the introduction, to achieve robustness, we need executives that make decisions dynamically and with low latency over expressive temporal representations. There are known methods for manipulating and reasoning over Simple Temporal Networks efficiently, which have been used as the foundation for most work in temporal executives. Furthermore, numerous efforts have formulated and developed extensions of STNs to include other useful properties, including uncertainty, preferences, and discrete choices. We briefly review some of these efforts. Since our work focuses on discrete choices, we discuss several efforts to build dynamic executives for these plans in more detail. These executives typically use one of two approaches: they reason on all plans in parallel, or switch between plans incrementally. However, these approaches, while promising, are typically either too memory intensive or may have high latency.

Temporal Constraint Networks (TCNs), formalized by Dechter, Meiri and Pearl (1991), capture many of the qualitative and metric temporal representations introduced by the AI community. A restricted type of TCN, the Simple Temporal Network, is used throughout recent work in temporal planning, temporal reasoning, and scheduling. Muscettola, Morris, and Tsamardinos (1998) proposed framework for low-latency dynamic execution: a pre-processing step called *compilation* and a run-time component called *dispatch*. Tsamardinos, Muscettola, and Morris (1998) later provided a faster compilation algorithm. Further work has also developed more efficient methods for testing consistency of STNs (Xu & Choueiry, 2003; Planken, de Weerdt, & van der Krogt, 2008).

Dechter et al. (1991) also proposed Temporal Constraint Satisfaction Problems, which include discrete choices that alter the simple interval constraint between particular pairs of events; each pair may have a choice of interval constraints, but choices for each pair of events must be independent. Stergiou and Koubarakis (2000) loosened this structural restriction, developing the Disjunctive Temporal Network (DTN). Tsamardinos, Pollack, and Ganchev (2001) presented the first dynamic executive for DTNs, which functions by generating the component STNs implied by all combinations of the discrete choices and compiling them independently, creating an exponential growth in memory use with respect to the number of choices.

Another important line of extension to STNs is the Simple Temporal Network with Uncertainty (STNU). Morris, Muscettola, and Vidal (2001) proved that an executive can test for consistency of an STNU and compile it into dispatchable form in polynomial time. Morris (2006) described a more efficient algorithm for testing dynamic controllability. Hunsberg (2009, 2010) corrected a flaw in the previous definitions and described an execution strategy using the more efficient dynamic controllability algorithm. Venable and Yorke-Smith (2005) added temporal uncertainty to DTNs. Tsamardinos (2002) introduced a probabilistic formulation of uncertainty in STNs. Conrad (2010) presents an extension of Drake to DTNs with uncertainty.

Tsamardinos, Vidal, and Pollack (2003) introduced Conditional Temporal Problems (CTP), adding uncontrollable discrete choices. The executive cannot control, but may only observe the values of some discrete choices at designated parts of the plan. Some of their notation is quite similar to that used here for Drake, but there are two important differences. First, a CTP is a strictly harder problem, since Drake is not concerned with uncontrollable choices, meaning that their algorithm does more work than is necessary for the simpler case. Second, their algorithm does not use a compact representation; their algorithm for consistency checking requires enumerating the possible scenarios. An open problem for future research is to adapt their more general algorithms to take advantage of the compactness of the Labeled Distance Graph.

Another useful feature added to STNs is preferences. Khatib, Morris, Morris, and Rossi (2001) introduced a formulation including preferences for event execution times within the simple interval bounds allowed by an STN, adding a notion of quality to the existing notion of consistency. Rossi, Venable, and Yorke-Smith (2006) discuss simultaneous handling of uncertainty and preferences.

Kim, Williams, and Abramson (2001) present Temporal Plan Networks, a representation that provides simple temporal constraints over durations combined in series, parallel, and with choice, where each choice has specified costs. Effinger (2006) expands this to a simple preference model, in which choices and activities have associated, fixed costs. Kirk is a dynamic executive for TPNs. Kirk performs optimal method selection just before run-time, assigning the discrete choices and then dispatches the resulting component STN. If some disturbance invalidates the STN that Kirk chose, then Kirk selects a new STN consistent with the execution thus far. Further research developed incremental techniques to allow Kirk to re-plan with lower latency (Shu, Effinger, & Williams, 2005; Block, Wehowsky, & Williams, 2006).

Shah and Williams (2008) present Chaski, an executive that dynamically dispatches plans with task assignment over heterogeneous, cooperative agents, represented by a TCN, by removing some redundant data structures and computations performed by Tsamardinos et al.'s (2001) algorithm. Shah and Williams point out that the component STNs of real-world TCNs often differ by only a few constraints, allowing a compact representation. They record all the component STNs by storing a single relaxed STN and maintaining a list of modifications to the relaxed STN that recover each of the original component STNs. By avoiding redundant records of shared constraints, their results show dramatic improvements in performance. Our work is inspired by the observation that this technique, although distinct, bears some resemblance to the environment labeling scheme that is employed by an Assumption Based Truth Maintenance System (ATMS). We specifically adapted ATMS ideas to work with a more general problem formulation than Chaski, expecting to see similar performance improvements.

Combi and Posenato (2009, 2010) discuss applications of dynamic executives to business work flows, which include flexibility over time of execution, hierarchical choice over execution paths, and temporal uncertainty. Their formalism for plans, *Workflow Schemata*, is closely related to DTN, STNU, and TPN frameworks, and they discuss variants of compilation and dispatching algorithms specialized to their representations. Their work describes an intriguing notion they call *history-dependent controllability*. Under this model, if event $X$ starts one of two sub-plans, the executive may not invalidate either sub-plan until it executes $X$ and begins one of them. Drake does not impose a similar requirement, but the requirement is certainly useful for preserving the executive's flexibility over future choices as the execution unfolds. Their algorithms for testing controllability enumerate the possible choices, and therefore suffers from memory growth.

Smith, Gallagher, and Zimmerman (2007) describe a distributed dynamic executive for real world plans in the C_TAEMS language. That representation uses STN temporal semantics and includes other features intended to represent cooperative multi-agent plans. Their language features a rich and practical notion of activity failure not present in STNs, including the potential for interruption of activities. The executive is given discrete choices over method selection and resource allocation, and attempts to maximize the utility of the overall plan. Their preference model accounts for partial or total method failures and supports different functions for accumulating reward, for example, summing or maxing. Their executive uses a re-planning strategy, similar to Kirk, which is enhanced by Hiatt, Zimmerman, Smith, and Simmons (2009) with a type of compile-time analysis called *strengthening*. This analysis performs a type of local repair that attempts to make the plan more robust to uncertainties or activity failures.

There are two central approaches to dynamic executives that include discrete choices. First, Tsamardinos et al.'s (2003) CTPs, Tsamardinos et al.'s (2001) DTN dispatcher, and Combi et al. (2010) use a compile-time analysis to compute the implied constraints of every possible plan and explicitly reason over them at run-time. Second, Kim et al. (2001), and Smith et al. (2007) focus on a single, potentially optimal, assignment to the choices, and if that becomes infeasible, they incrementally re-plan and extract a new plan. Both methods have shortcomings, since explicit compilation is memory intensive and re-planning steps can be computationally intensive, especially if the executive is forced to re-plan often. Drake, like Chaski, provides a middle ground by working with a compilation strategy that has a reduced memory footprint.

## 2.2 Background on ATMSs

Stallman and Sussman (1977) introduced the profoundly useful idea of tracing the dependency of deductions in a computerized aid to circuit diagnosis, in order to focus the search for a consistent component mode assignment during transistor circuit analysis. The dependencies it computes from solving the equations allow it to rapidly find those choices that might be responsible for the detected failure. They generalize this approach to combinatorial search, introducing the dependency-directed backtracking algorithm, which ensures that when a conflict is found that the search backs up far enough to ensure that the newly found inconsistency is actually removed.

Doyle (1979) introduced Truth Maintenance Systems (TMSs) as a domain independent method for supporting dependency-directed backtracking. The TMS represents data, their justifications, and provides the ability to revise beliefs when assumptions change or contradictions arise. For example, consider a problem solver designed to search for a solution to a constraint satisfaction problem. When determining whether a particular solution is consistent, the problem solver will perform a chain of inferences, providing the TMS with the justification for each step. If an inconsistency is found, the problem solver selects a new candidate solution, and the TMS uses the justifications to determine which of the inferences still hold under the new candidate and which must be recomputed to account for new circumstances. The TMS continually determines whether a particular *datum*, a general term for any fact that arises in problem solving, is *in* or *out*, that is, currently believed true, or not currently believed.

Later work relaxes the goal of maintaining a single, consistent assignment to all data of in or out, and instead tracks the contexts in which particular facts hold, even if those contexts may be mutually exclusive. McDermott (1983) uses *beads* to state a context, which is a particular set of choices or assumptions on which the reasoning might rely, and provides *data pools* that specify all the facts that hold in that context. De Kleer (1986) develops the Assumption-based Truth Maintenance System, which uses a similar idea, but changes the terminology to use *environments* and *labels* to specify contexts. The ATMS maintains a set of minimal inconsistent environments, called conflicts or no-goods. These conflicts help the system to avoid performing inferences for contexts that are already known to be inconsistent, and the minimality of the conflict set makes the procedure tractable. The ATMS is designed to simultaneously find the logical consequences of all possible combinations of assumptions, in contrast to the TMS, which focuses on finding any one set of assumptions that solve the problem of interest. Hence, the ATMS is well suited as the foundation for an executive that is intended to consider all possible choices simultaneously without incurring latency for switching between choices. Finally, there is some development of the idea that when working with inequalities, the ATMS only needs to keep tightest bounds on the inequalities, which we use extensively; this concept was described by Goldstone (1991) as *hibernation*.

We leave our review of the details of the ATMS to later sections, as we develop Drake's machinery in depth.

## 3. Dynamic Execution of STNs and Labeled STNs

Now we are prepared to present the formal representation of the temporal plans that Drake uses. Plans are composed of actions that need to be performed at feasible times, where we define feasible times through constraints on the start and end times of the plan activities. Our work builds upon Simple Temporal Networks, and we begin by explaining how STNs constrain the events of a plan and their feasible execution. We extend these definitions to include discrete choices, thereby constructing Labeled STNs.

## 3.1 Simple Temporal Networks

Simple Temporal Networks provide a framework for efficiently reasoning about a limited form of temporal constraints. A simple temporal network is defined as a set of events related by binary interval constraints, called simple interval constraints (Dechter et al., 1991).

**Definition 3.1 (Event)** An event is a real-valued variable, whose value is the execution time of the event. □

**Definition 3.2 (Simple Interval Constraint)** A simple interval constraint $\langle A, B, l, u \rangle$ between two events $A$ and $B$ requires that $l \leq B - A \leq u$, denoted, $[l, u]$. □

By convention, $u$ is non-negative. The lower bound, $l$ may be positive if there is a strict ordering of the events, or negative if there is not a strict ordering. Positive or negative infinities may be used in the bounds to represent an unconstrained relationship.

**Definition 3.3 (Simple Temporal Network)** A Simple Temporal Network $\langle V, C \rangle$ is comprised of a set of events $V$ and a set of simple interval constraints $C$. A *schedule* for an STN is an assignment of a real number to each event in $V$, representing the time to schedule each event. The Simple Temporal Problem (STP) is, given an STN $\langle V, C \rangle$, return a consistent schedule if possible, else return false. A *consistent schedule* is a schedule that satisfies every constraint in $C$. If and only if at least one solution exists, the STN is *consistent*. □

**Definition 3.4 (Dynamic Execution)** A *dynamic execution* of an STN is the construction of a consistent schedule for the STN in real-time. The executive decides at time $t - \epsilon$ whether to execute any events at time $t$, for some suitably small $\epsilon$. If at some time, there are no longer any remaining consistent schedules, return false immediately. The executive may arbitrarily select any consistent schedule. □

## 3.2 Adding Choice: Labeled STNs

This section defines our key representational concept, Labeled STNs, a variant of STNs that is designed to include discrete choices between temporal constraints. Although equivalent in expressiveness to Disjunctive Temporal Networks, Labeled STNs provide more consistent terminology for Drake, and their corresponding labeled distance graphs make it easier to extend standard STN and weighted graph algorithms to include choice. We show the precise connection between DTNs and Labeled STNs at the end of this section. These definitions are used throughout the compilation and dispatching algorithms presented later in this work.

The input problem needs a succinct way to state what the choices of the input plan are and what the possible options are that the executive may select between. We accomplish this through a set of finite domain variables.

**Definition 3.5 (Choice Variables)** Each choice is associated with a finite domain variable $x_i$. Each of these variables has a domain with size equal to the number of options of that choice. $X$ is the set of all the variables for a particular problem. An *assignment* is a selection of a single option for a choice, represented as an assignment to the choice's associated choice variable. □

**Example 3.6** In the rover example, there is a single choice with two options, leading to a single variable $x \in \{collect, drive\}$. We might assign $x = collect$ to represent the choice of collecting samples. □

In general, Drake will reason about the implications of combinations of assignments. To specify these assignments to the choice variables, Drake uses *environments*. Following the ATMS, Drake annotates, or "labels" interval constraints and edge weights with environments specifying when they are entailed.

**Definition 3.7 (Environment)** An *environment* is a partial assignment of the choice variables in $X$, written $e = \{x_i = d_{ij}, ...\}$. An environment may have at most one assignment to each variable to be consistent. A *complete environment* contains an assignment to every choice variable in $X$. An *empty environment* provides no assignments and is written $\{\}$. We denote the set of possible environments as $\mathcal{E}$ and the set of complete environments as $\mathcal{E}_c$. The length of an environment is the number of assigned variables, denoted $|e|$. □

**Example 3.8** In a problem with two choice variables $x, y \in \{1, 2\}$, some possible environments are $\{\}$, $\{x = 1\}$, $\{y = 2\}$, and $\{x = 1, y = 2\}$. □

In a Labeled STN, different assignments to the choice variables entail different temporal constraints, which we represent with labeled simple interval constraints.

**Definition 3.9 (Labeled Simple Interval Constraint)** A labeled simple interval constraint is a tuple $\langle A, B, l, u, e \rangle$ for a pair of events $A$ and $B$, real valued weights $l$ and $u$ and environment $e \in \mathcal{E}$. This constraint states that, if the assignments in $e$ hold, then the simple interval constraint $\langle A, B, l, u \rangle$ is entailed. □

Then a Labeled STN is defined analogously to an STN, but extended to include choices and labeled constraints.

**Definition 3.10 (Labeled Simple Temporal Network)** A Labeled STN $\langle V, X, C \rangle$ is a set of events $V$, a set of choice variables $X$, and a set of labeled simple interval constraints $C$. As with STNs, a *schedule* for a Labeled STN is an assignment of real numbers to each event, indicating the time to execute the event. This schedule is *consistent* if there is a full assignment to the choice variables so that the schedule satisfies every simple interval constraint entailed by a labeled simple interval constraint. □

**Example 3.11 (Rover Problem as a Labeled STN)** The rover problem of Example 1.1 has a single choice, with two options, collecting samples or charging the batteries. Use a single choice variable $x \in \{collect, charge\}$ to represent the choice. The network has events $\{A, B, C, D, E, F\}$. The labeled simple interval constraints below are written $e : l \leq B - A \leq u$.

$$\{\} : 0 \le F - A \le 100 \tag{1}$$

$$\{\} : 30 \le B - A \le 70 \tag{2}$$

$$\{\} : 0 \le F - E \le 0 \tag{3}$$

$$\{x = collect\} : 50 \le C - B \le 60 \tag{4}$$

$$\{x = collect\} : 0 \le E - C \le 0 \tag{5}$$

$$\{x = charge\} : 0 \le D - B \le 50 \tag{6}$$

$$\{x = charge\} : 0 \le E - D \le 0 \tag{7}$$

This notation states that the inequalities on lines 1-3 must always hold, lines 4-5 must hold if the executive decides to collect samples, and lines 6-7 must hold if the executive decides to charge the batteries. The schedule given in Example 1.2, $A = 0, B = 45, C = 95, D = 45, E = 95, F = 95$, is consistent with the full assignment $x = collect$. The constraints on lines 1-3 hold for any choice, so necessarily must hold here. Lines 4-5 give constraints with environments whose assignments are all given in the full assignment, so their simple interval constraints must hold, and do. Lines 6-7 give constraints with an environment that include different assignments than our full assignment, so the constraints do not need to hold for the schedule to be consistent. □

Drake provides a dynamic execution of a Labeled STN, making decisions at run-time, as late as possible.

**Definition 3.12 (Dynamic execution of a Labeled STN)** A dynamic execution of a Labeled STN is the simultaneous real-time construction of a full assignment and a corresponding consistent schedule. The executive decides at time $t - \epsilon$ whether it will execute any events at time $t$, for some suitably small $\epsilon$. Possible assignments to choices are only eliminated from consideration as necessary to schedule events. The executive may arbitrarily select any consistent schedule. □

Example 1.2 gave a dynamic execution of the rover problem because the dispatcher selected choices and scheduling times dynamically, only choosing the option to collect samples immediately before the start of the sample collection activity.

We conclude this section by discussing the equivalence of DTNs and Labeled STNs, to allow for easier comparison to prior work. Recall the definition of DTNs (Stergiou & Koubarakis, 2000).

**Definition 3.13 (Disjunctive Temporal Network)** A Disjunctive Temporal Network $\langle V, C \rangle$ is a set of events $V$ and a set of disjunctive constraints $C$. Each disjunctive constraint $C_i \in C$ is of the form

$$c_{i1} \vee c_{i2} \vee ... \vee c_{in}, \tag{8}$$

for positive integer $n$ and each $c_{ij}$ is a simple interval constraint. As before, a schedule is an assignment of a time to each event. The schedule is consistent if and only if at least one simple interval constraint $c_{ij}$ is satisfied for every disjunctive constraint $C_i$. A DTN is *consistent* if and only if at least one consistent schedule exists. □

The DTN and Labeled STN definitions are analogous, except for the difference in how choices are specified. We can construct a Labeled STN equivalent to any DTN by creating a choice variable for each disjunctive constraint, with one value for each disjunct. Thus, $x_i = 1...n$. Each disjunctive constraint in the DTN, $c_{ij}$ is labeled with environment $x_i = j$. Non-disjunctive constraints are labeled with $\{\}$.

**Example 3.14** Consider a DTN with three events, $A, B, C$. Assume there are two disjunctive constraints, $\langle A, B, 3, 5 \rangle$, and $\langle B, C, 0, 6 \rangle \vee \langle A, C, -4, -4 \rangle$. Then the corresponding Labeled STN would have one binary choice, represented by $x \in \{1, 2\}$. It would have three labeled simple interval constraints: $\langle A, B, 3, 5, \{\} \rangle$, $\langle B, C, 0, 6, \{x = 1\} \rangle$, and $\langle A, C, -4, -4, \{x = 2\} \rangle$. □

To see the reverse construction, note that the DTN specifies a set of simple interval constraint in conjunctive normal form. Labeled STNs allow the specification of somewhat more complex boolean expressions, but all boolean expressions are reducible to conjunctive normal form, so Labeled STNs are not more expressive. The mapping from DTNs to Labeled STNs is straightforward because we construct a Labeled STN that directly uses a conjunctive normal form expression.

**Example 3.15** Consider the rover problem, given as a Labeled STN in Example 3.11. If we represent the constraint $0 \leq F - A \leq 100$ as $C_{F,A}$, then we can identify that this problem has the following boolean form

$$C_{F,A} \wedge C_{B,A} \wedge C_{F,E} \wedge ((C_{C,B} \wedge C_{E,C}) \vee (C_{D,B} \wedge C_{E,D})) \tag{9}$$

We could expand this to conjunctive normal form

$$C_{F,A} \wedge C_{B,A} \wedge C_{F,E} \wedge (C_{C,B} \vee C_{D,B}) \wedge (C_{C,B} \vee C_{E,D}) \wedge (C_{D,B} \vee C_{E,C}) \tag{10}$$

From this form, we can construct a DTN directly. □

Thus, like the DTN, Labeled STNs provide a rich notion of choice. Given the definition of the problem that Drake solves in this section and the introduction of environments, the next two sections develop the labeled machinery Drake uses to efficiently perform temporal reasoning.

## 4. Distance Graphs and Temporal Reasoning

Dechter et al. (1991) showed that STN reasoning can be reformulated as a shortest path problem on an associated weighted distance graph. This connection is important because weighted graphs are easy to manipulate and have well developed theory and efficient algorithms, hence most practical algorithms for STNs are based on this connection. Drake follows the prior literature, in that it frames the temporal reasoning as a labeled version of shortest path problems. This section begins to develop the formalism for Labeled Value Sets and Labeled Distance Graphs, which allow us to compactly represent the shortest path problems and algorithms. We begin by reviewing the transformation for STNs.

**Definition 4.1 (Distance Graph associated with an STN)** A distance graph associated with an STN is a pair $\langle V, W \rangle$ of vertices $V$ and edge weights $W$. Each event is associated with a vertex in $V$. The vertices exactly correspond to the events of the STN. The weights are a function $V \times V \to \mathbb{R}$. The simple temporal constraint $l \leq B - A \leq u$ is represented by the edge weights $W(B, A) = u$ and $W(A, B) = -l$. □

Figure 1.2 shows an example of the conversion from an STN to a distance graph. Recall that an STN is consistent if and only if its associated distance graph does not have any negative cycles (Dechter et al., 1991). The compilation algorithm for an STN takes the associated distance graph as an input and outputs another distance graph that is the dispatchable form.

**Definition 4.2 (Dispatchable form of a distance graph)** A weighted distance graph is the dispatchable form of an STN if an executive may dynamically execute the STN in a greedy fashion to construct a consistent schedule using only local propagations. The dispatchable form is *minimal* if all the edges are actually needed by the dispatcher for correct execution. □

The all-pairs shortest path (APSP) graph of the distance graph associated with an STN is a dispatchable form because it explicitly contains all possible constraints of the STN (Muscettola et al., 1998). The minimal form is computable by performing pruning on the APSP graph.

We begin building the labeling formalism by defining *labeled value pairs*. Just as we labeled simple temporal constraints with environments, labeled values associate values with environments.

**Definition 4.3 (Labeled Value Pair)** A *Labeled Value Pair* is a pair $(a, e)$, where $a$ is a value and $e \in \mathcal{E}$ is an environment. The value $a$ is entailed (usually as an assignment or an inequality bound) under all environments where the assignments in $e$ hold. □

During compilation and dispatch, Drake uses labeled value pairs to track real valued bounds, so $a \in \mathbb{R}$. During compilation, when performing shortest path computations, Drake tracks predecessor vertices, so the value may be a pair $(b, v) \in \mathbb{R} \times V$. Having these two types of values complicates our discussion somewhat, but is necessary for the compilation algorithms, and allows an elegant implementation of the relaxation algorithm for directed, weighted graphs. The ATMS strategy of associating environments with values is well founded for any arbitrary type of value, so both choices are sound. Labeled value pairs always use minimal environments, that is, the environment specifies the smallest set of assignments possible for the implication to be true. This minimality is critical for the labeling system to be efficient.

**Example 4.4** If there is a choice variable $x \in \{1, 2\}$, then $(3, \{x = 1\})$ is a real valued labeled value pair. Similarly, if $A$ is an event, $((2, A), \{x = 2\})$ is a possible predecessor graph labeled value pair. □

Now we have arrived at a crucial contribution of this paper: extending domination into the labeled space. If we have the inequality $B - A \leq 4 \leq 5$, then it is clear that we only need to keep the dominant value, the four, and may discard the five. Shortest path algorithms widely use the concept of dominance to propose many possible paths and keep only the tightest one. When applied to labeled value pairs, dominance involves an ordering on two parts, the value and the environment. In Drake, values are always ordered using real valued inequalities, either $\leq$ or $\geq$. This paper mostly uses $\leq$, except when reasoning on lower bounds during dispatch, where explicitly noted. When necessary, a direct replacement of the inequality direction suffices to extend the definitions. Environments are ordered through the concept of subsumption.

**Definition 4.5 (Subsumption of Environments)** An environment $e$ subsumes $e'$ if for every assignment $x_i = d_{ij} \in e$, the same assignment exists in $e'$, denoted $x_i = d_{ij} \in e'$. □

**Example 4.6** An environment $\{x = 1, y = 2, z = 1\}$ is subsumed by $\{x = 1, z = 1\}$ because the assignments in the later environment are all included in the former. □

Then domination of labeled value pairs applies to both orderings simultaneously.

**Definition 4.7 (Dominated Labeled Value Pair)** Let $(a, e_a), (b, e_b)$ be two labeled value pairs where the values are ordered with $\leq$. Then $(a, e_a)$ dominates $(b, e_b)$ if $a \leq b$ and $e_a$ subsumes $e_b$. □

**Example 4.8** If $B - A \leq 4$ under $\{\}$ and also $B - A \leq 4$ under $\{x = 1\}$, then the first inequality is non-dominated because its environment is less restrictive. Thus $(4, \{\})$ dominates $(4, \{x = 1\})$. Likewise, if $B - A \leq 2$ under $\{x = 1\}$, this constraint dominates the constraint $B - A \leq 4$ under $\{x = 1\}$, because the first inequality is tighter and holds within every environment where the second inequality holds, hence $(2, \{x = 1\})$ dominates $(4, \{x = 1\})$. □

To represent the various values a bound may have, depending on the choices the executive makes, Drake collects the non-dominated labeled value pairs into a *Labeled Value Set*.

**Definition 4.9 (Labeled Value Set)** A *Labeled Value Set*, $L$ is a set of non-dominated labeled value pairs, that is, a set of pairs $(a_i, e_i)$. Thus, we may write $L = \{(a_1, e_1), \ldots (a_n, e_n)\}$. No labeled value pair in the set is dominant over another pair in the set. □

In any particular labeled value set, the values are all of the same type, either real values or value/vertex pairs.

**Example 4.10** Assume there is a single choice variable, $x \in \{1, 2\}$. Some real valued variable, $A \in \mathbb{R}$ has value 3 if $x = 1$ and 5 if $x = 2$. Then A is represented by the labeled value set, $A = \{(3, \{x = 1\}), (5, \{x = 2\})\}$. □

Finally, we can modify distance graphs to use Labeled Value Sets instead of real values for the weights, leading to the definition of labeled distance graphs.

**Definition 4.11 (Labeled Distance Graph)** A labeled distance graph $G$ is a tuple $\langle V, E, W, X \rangle$. $V$ is a set of vertices and $E$ is a set of directed edges between the vertices, represented as ordered pairs $(i, j) \in V \times V$. $W$ is a weight function from edges to real valued labeled value sets, where each edge $(i, j) \in E$ is associated with a labeled value set $W(i, j)$. $X$ is the description of the choices variables, defining the set of environments that may appear in the labeled value sets contained in $W$. $\square$

**Example 4.12** Consider a simple graph with three vertices, $V = \{A, B, C\}$. This graph contains edges $E = \{(A, B), (A, C), (B, C)\}$. There is one choice variable $x \in \{1, 2\}$. Edge $(A, B)$ has weight 1 regardless of the choice. Edge $(A, C)$ is 3 if $x = 1$ and 7 if $x = 2$. Finally, edge $(B, C)$ has weight 4 if $x = 2$. This labeled distance graph is shown in Figure 4.1. $\square$

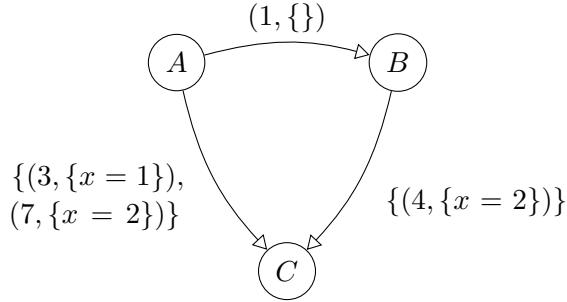

Figure 4.1: A simple Labeled Distance Graph with one choice variable, $x \in \{1, 2\}$

The association between Labeled STNs and labeled distance graphs closely parallels the association between STNs and distance graphs.

**Definition 4.13 (Labeled Distance Graph Associated with a Labeled STN)** The labeled distance graph associated with a Labeled STN has a vertex associated with every event in the Labeled STN. For each labeled inequality, denoted $e : Y - X \leq w$, where $e \in \mathcal{E}$, $X, Y \in V$, $w \in \mathbb{R}$, and the edge $(X, Y) \in E$. The labeled value set $W(X, Y)$ includes the labeled value pair $(w, e)$ if the pair is not dominated by another $(w', e') \in W(X, Y)$. $\square$

**Example 4.14** Figure 1.4 is the essentially the labeled distance graph for the rover problem described in Example 1.1, except that we replace the informal notation with equivalent labeled value sets. For example, the edge $(B, C)$ currently has weight $S : 60$, which we replace with the labeled value set $\{(60, \{x = collect\})\}$. $\square$

Drake compiles the input Labeled STN by creating its associated labeled distance graph representation and then constructing a new labeled distance graph through a transformation of the input graph.

**Definition 4.15 (Dispatchable Form of a Labeled STN)** A labeled distance graph is the dispatchable form of a Labeled STN if it represents all the constraints necessary to accurately perform dispatch with a greedy strategy, using only local propagations. $\square$

This section constructed the labeling structures Drakes uses, beginning with labeled value pairs and then building labeled value sets from minimal sets of labeled pairs. Labeled value sets are used to construct labeled distance graphs, and we defined how to construct the graph associated with a Labeled STN, on which Drake's compilation algorithms operate.

## 5. Labeled Value Set Maintenance System

The previous section defines the representation for labeled distance graphs and labeled value sets; this section provides tools for manipulating them. The primary focus of labeled value sets is to maintain only non-dominated labeled value pairs, which makes the labeled value sets compact and efficient. This section introduces three concepts. First, we describe how to extract values from a labeled value set, called a *query*, allowing us to find the dominant value implied by an environment. Second, how to handle assignments to choices that are inconsistent, which are called *conflicts*. Third, how to apply functions to labeled value sets, which allows us to perform computations on them directly.

First we define the query, which lets us extract the precise dominant value(s) that are guaranteed to hold under a particular environment.

**Definition 5.1 (Labeled Value Set Query)** The query operator $A(e)$ is defined for labeled value set $A$ and $e \in \mathcal{E}$. The query returns a set of values including $a_i$ from the pair $(a_i, e_i) \in A$, where $e_i$ subsumes $e$, if and only if there is no other pair $(a_j, e_j) \in A$ such that $e_j$ subsumes $e$ and $a_j < a_i$. If no environment $e_i$ subsumes $e$, then $A(e)$ returns $\emptyset$. □

Here we introduce a convention: when using pairs of real values and vertices for the values of labeled value pairs, we consider inequalities over the pairs to be defined entirely by the real value. Thus, $(5, A) < (6, E)$ and $(5, A) \not< (5, E)$.

**Example 5.2** Assume that there are two choice variables $x, y \in \{1, 2\}$, and a real valued variable $A$ represented by a labeled value set, where $A = \{(3, \{x = 1\}), (5, \{\})\}$. $A(\{x = 1, y = 1\}) = 3$ because the input environment is subsumed by both environments in the labeled value set, but three is dominant. $A(\{x = 2, y = 2\}) = 5$ because only the empty environment in the labeled value set subsumes the input environment. □

**Example 5.3** Let $A$ be a labeled value set with values that are pairs of real values and vertices, $A = \{((3, A), \{x = 1\}), ((5, A), \{y = 1\}), ((5, B), \{\})\}$. Then $A(\{x = 1, y = 2\}) = \{(3, A)\}$ because all the environments are subsumed, but $(3, A) < (5, A)$ and $(3, A) < (5, B)$. $A(\{x = 2, y = 1\}) = \{(5, A), (5, B)\}$ because the tighter value is not applicable and neither value with real part five is less than the other. □

The query operator defines the expansion from the compact form to an explicit listing of the values for each environment. Although Drake never performs this expansion, it is useful for determining the correct behavior of a Labeled Value Set when designing algorithms.

Second, we consider *conflicts*, an important function of an ATMS that allows it to track inconsistent environments. In our case, an inconsistent environment signals an inconsistent component STN. The standard strategy in an ATMS is to keep a list of minimal *conflicts*, also referred to as *no-goods* (de Kleer, 1986).

**Definition 5.4 (Conflict)** A *conflict* is an environment $e$ such that if $e$ subsumes $e'$, then $e'$ is inconsistent. The conflict $e$ is *minimal* if there is no $e'' \in \mathcal{E}$ such that $e'' \subset e$, such that $e''$ is a conflict (Williams & Ragno, 2007). □

**Example 5.5** For example, the compilation process might determine that $x = 1$ and $y = 1$ are contradictory choices, and cannot be selected together during any execution. Then, $\{x = 1, y = 1\}$ is a conflict. □

Often, reasoning algorithms keep a cache of conflicts to avoid performing work on environments that were previously discovered to contain an inconsistency. In practice, the set of conflicts can become large and unwieldy, leading some practical systems to keep only a subset of conflicts, using principles such as temporal locality to maintain a small, useful cache. Since the cache of conflicts is incomplete, the cache can miss, requiring the problem solver to re-derive an inconsistency, but missing the cache would never lead to incorrectly accepting an inconsistent solution. A good example of this is conflict learning in the widely studied SAT-solver MiniSAT (En & Srensson, 2004). During real-time execution, an incomplete cache of conflicts would require Drake to perform non-local propagation to re-test for inconsistency in the case of a cache miss. This extra step violates the principles of dispatchable execution. Therefore, Drake maintains a complete cache of all known conflicts, allowing Drake to verify that an environment is not known to be inconsistent with a single check of the cache. Furthermore, Drake can quickly test whether any complete environments are valid, as the inconsistencies are all readily available.

If there is a known conflict, Drake sometimes needs to determine how to *avoid* a conflict, that is, what minimal environments ensure the conflict is not possible.

**Definition 5.6 (Constituent Kernels, Williams et al., 2007)** A conflict $e_c$ has an associated set of constituent kernels, each of which is an environment that specifies a single assignment that takes some variable assigned in the conflict and assigns it a different value. Hence, if $e_k$ is a constituent kernel, then any other environment $e$ such that $e_k$ subsumes $e$ implies that $e_c$ does not subsume $e$, and hence is not subject to that conflict. Thus, we say $e$ avoids the conflict. □

**Example 5.7** If there are three variables, $X, Y, Z \in \{1, 2\}$, assume $\{x = 1, y = 1\}$ is a conflict. Then the constituent kernels are $\{x = 2\}$ and $\{y = 2\}$, as any complete environment that does not contain the conflict must assign either $x$ or $y$ not to be one. □

The final tool needed is the ability to perform temporal and graph reasoning on labeled value sets, which is principally accomplished through computing path lengths or propagating inequality bounds. We can construct the approach by denoting the inference rule on values as a function $f$. Then, we can build the method for applying $f$ to labeled value sets from the rule for applying it to labeled value pairs. We begin by defining the union operation on environments, which is the fundamental operation on environments during temporal reasoning, as in the ATMS literature (de Kleer, 1986).

**Definition 5.8 (Union of Environments)** The *union* of environments, denoted $e \cup e'$ is the union of all the assignments of both environments. If $e$ and $e'$ assign different values to some variable $x_i$, then there is no valid union and $e \cup e' = \bot$, where $\bot$ is the symbol for false. If $e \cup e'$ is subsumed by a conflict, then $e \cup e' = \bot$. This value signifies that there is no consistent environment where both $e$ and $e'$ hold simultaneously. □

**Example 5.9** Most commonly, unions are used to compute the dependence of new derived values. If $A = 2$ when $\{x = 1\}$ and $B = 3$ when $\{y = 2\}$, then $C = A + B = 5$ when $\{x = 1\} \cup \{y = 2\} = \{x = 1, y = 2\}$. □

Using this notation, then an inference on labeled value pairs involves performing that inference on the values to produce a new value, and unioning the environments to produce a new environment.

**Lemma 5.10** Consider labeled value pairs $(a, e_a), (b, e_b)$. Applying some function $f$ to the pair yields $(f(a, b), e_a \cup e_b)$. □

**PROOF** For any environment $e$, if $e_a$ subsumes $e$ then $a$ is entailed, and if $e_b$ subsumes $e$ then $b$ is entailed, by the definition of labeled value pairs. Additionally, $e_a \cup e_b$ subsumes $e$ if and only if $e_a$ subsumes $e$ and $e_b$ subsumes $e$. Therefore, $e_a \cup e_b$ subsumes $e$ entails both values $a, b$, and hence allows us to compute $f(a, b)$. Thus, $(f(a, b), e_a \cup e_b)$ is well defined. □

Note that in this lemma, $e_a \cup e_b$ may produce $\bot$, which would indicate that the labeled value pair never holds in a consistent environment and may be discarded.

**Example 5.11** Consider computing the quantity $(3, \{x = 1\}) + (6, \{y = 1\}) = (3 + 6, \{x = 1\} \cup \{y = 1\}) = (9, \{x = 1, y = 1\})$. □

If the function respects the dominance of values, then we may apply it to entire labeled value sets.

**Definition 5.12** A function $f(a, b)$ is consistent with $\leq$ dominance if for values $a, b, c$ and $d$, $a \leq c, b \leq d \implies f(a, b) \leq f(a, d), f(a, b) \leq f(c, b), f(a, b) \leq f(c, d)$. □

**Lemma 5.13** If variables $A, B$ are represented by labeled value sets and $f$ is consistent with the domination ordering used for $f$, then the result $C = f(A, B)$ is represented by a labeled value set containing a non-dominated subset of the labeled value pairs $(f(a_i, b_j), e_i \cup e_j)$ constructed from every pair of labeled value pairs $(a_i, e_i) \in A$ and $(b_j, e_j) \in B$. □

**PROOF** The lemma follows from the argument that the cross product of applying $f$ to all the labeled value pairs in $A$ and $B$ produces all possible labeled value pairs. Since we must ensure that the labeled value pairs include all the dominant labeled value pairs for $C$, we require $f$ to be consistent with the ordering so that the dominant values of $A$ and $B$, which are all we have available, are sufficient to derive the dominant labeled value pairs of $C$. □

We conclude this section with an example of relaxation of a labeled distance graph, which is a core rule of inference for most weighted graph algorithms that uses the path $A \rightarrow B \rightarrow C$ to compute a possible path weight $A \rightarrow C$. The derived path length replaces the old path if the newly derived path length is shorter.

**Example 5.14** Consider the labeled distance graph in Figure 4.1. We compute $W(A, B) + W(B, C) = \{(1, \{\})\} + \{(4, \{x = 2\})\} = \{(5, \{x = 2\})\}$. Compare this to the existing known weights $W(A, C) = \{(3, \{x = 1\}), (7, \{x = 2\})\}$, and we determine that the new value replaces the old value of 7. After the update, $W(A, C) = \{(3, \{x = 1\}), (5, \{x = 2\})\}$. □

There are two fundamental inferences performed with labeled value sets: relaxation of weighted edges during compilation, and sums and differences of edge weights to compute bounds on execution times during dispatch. Both follow the framework outlined here, and are explained in more detail in the compilation and dispatch sections, respectively. These operations complete the definition of labeled value sets and the following sections use them to construct compact compilation and dispatching algorithms.

## 6. Dispatching Plans with Choice

Given this foundation in Labeled STNs, labeled value sets, and labeled distance graphs, we turn to the central focus of this article - dynamic execution of Labeled STNs. Recall that the dispatcher uses a local, greedy algorithm to make decisions at run-time with low latency, and that the accuracy of this approach is guaranteed by the compilation step. We begin with the dispatcher because low latency execution is the fundamental goal of this work and because, as in prior work, the compiler is designed to produce an output appropriate for this dispatcher.

We adapt the STN dispatcher developed by Muscettola et al. (1998) to work with dispatchable labeled distance graphs. Essentially, Drake's algorithms substitute real number bounds on execution times, as in the STN dispatcher, with labeled value sets. Additionally, we adapt Tsamardinos et al.'s (2001) approach to reasoning over multiple possible options, that is, allowing the dispatcher to accept a proposed execution time for an event if at least one full assignment to the choices is consistent with that schedule. We present Drake's dispatching algorithms by first reviewing standard STN dispatching, and then adapt these techniques to handle labels.

### 6.1 STN Dispatching

Muscettola et al. (1998) showed that given a dispatchable form of the STN, a simple greedy dispatcher can correctly execute the network with updates performed only on neighboring events in the dispatchable form of the STN. The dispatcher loops over the non-executed events at each time step, selecting an event to execute if possible, or else waiting until the next time step. This process continues until either all the events are executed or a failure is detected.

Determining whether an event is executable relies on two tests. First, the dispatcher tests whether the ordering constraints of an event have been satisfied, which is called testing for *enablement*. A simple temporal constraint may imply a strict ordering between two events, which the dispatcher must explicitly test to ensure that an event is not scheduled before an event that must precede it. Second, the dispatcher efficiently tracks the consequences of the simple temporal constraints between the event and its neighbors by computing *execution windows* for each event. Execution windows are the tightest upper and lower bounds derived for each event through one-step propagations of execution times. If the current time is in an event's execution window and it is enabled, the event may be scheduled at the current time.

Now we briefly recall the derivation of these two rules for executing events. Recall that each weighted edge in the distance graph corresponds to an inequality

$$B - A \leq w_{AB} \tag{11}$$

where $A$ and $B$ are execution times of events and $l$ is some real number bound. If we select execution time $t_A$ for event $A$ and $B$ is not yet scheduled, then we can rearrange the inequality as

$$B \leq w_{AB} + t_A \tag{12}$$

This produces a new upper bound for the execution time of $B$. Likewise, if $A$ is not yet scheduled and we select $t_B$ as the execution time of $B$, then we can rearrange the inequality as

$$A \geq t_B - w_{AB}. \tag{13}$$

Thus, we derive a new lower bound for $A$. If, in this form, $w_{AB} < 0$, then $B < A$, so event $B$ must precede $A$, implying an enablement constraint.

We can recast these rules in terms of propagations on the distance graph. If an event $A$ is scheduled at time $t$, then propagate it through all outbound edges $(A, B)$ to derive upper bounds $B \leq w_{AB} + t$, and through all inbound edges $(B, A)$ to derive lower bounds $B \geq t - w_{BA}$. Event $B$ has event $A$ as a predecessor if there is a negative weight edge $(B, A)$. As usual, dispatching is only affected by the dominant upper and lower bounds, so the dispatcher only stores the dominant constraints.

**Example 6.1** Consider the dispatchable distance graph fragment in Figure 6.1. The execution windows begin without constraint, $-\infty \leq A \leq \infty$ and $-\infty \leq B \leq \infty$. If we begin execution at $t = 0$, $B$ is not executable yet because it is not enabled; the negative weighted edge $(B, A)$ implies that $A$ must be executed first. $A$ has no predecessor constraints and zero lies within its execution bound, so may be executed at this time. Propagating the time $A = 0$ allows us to derive the bounds $2 \leq B \leq 8$. Then $B$ may be executed at any time between 2 and 8. If the dispatcher reached time 9 without having executed $B$, then it must signal a failure. □

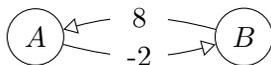

Figure 6.1: A dispatchable distance graph fragment.

The final consideration for the dispatcher is zero-related events. If two events are constrained to be executed at precisely the same time, then the prior work requires that the events are collapsed into a single vertex in the dispatchable graph. Otherwise, zero-related vertices may cause the dispatcher to make mistakes because they must occur together, yet they appear independently schedulable. Equivalently, the dispatcher may simulate the collapse by always executing zero-related vertices as a set. For example, if $A$ and $B$ are known to be zero-related, the dispatcher may schedule them together, scheduling the events at time $t$ if the execution windows and enablement constraints of both $A$ and $B$ are satisfied. Note that since they are zero-related, it is impossible to have an enablement constraint between them.

## 6.2 Labeled STN Dispatching

Drake relies on these same fundamental structures and rules as the STN dispatcher, but modifies each step to consider labels. Since edge weights are labeled value sets, the execution windows are also labeled value sets, implying that the upper and lower bounds of execution times for events may vary depending on the assignments to choices and may vary separately. For each possible bound or enablement constraint, Drake's dispatcher must either enforce the constraint, or discard the constraint and decide not to select the associated environment. Broadly, this is the same strategy as Tsamardinos et al. (2001), where the component STNs are dispatched in parallel, and proposed scheduling decisions may be accepted if they are consistent with at least one STN.

We begin by considering how to update the propagation rules to derive execution windows. The STN propagations involve adding edge weights (or negative weights) to the execution time of an event. The upper bound of every event is initially $\{(\infty, \{\})\}$ and every lower bound is initially $\{(-\infty, \{\})\}$. Upper bounds are dominated by low values, or the $\leq$ inequality, and lower bounds are dominated by large values, or the $\geq$ inequality. The following theorem describes the propagation of execution bounds, which is the labeled implementation of Equations 12 and 13.

**Theorem 6.2** *If event $A$ is executed at time $t$, then consider some other event $B$. For every $(w_{AB}, e_{AB}) \in W(A, B)$, $(w_{AB} + t, e_{AB})$ is a valid upper bound for execution times of $B$ and for every $(w_{BA}, e_{BA}) \in W(B, A)$, $(t - w_{BA}, e_{BA})$ is a lower bound for execution times of $B$.* □

PROOF These rules are a direct extension of the STN propagation to the labeled case using labeled operations as in Definition 5.12. Since the execution actually occurred, it holds under all possible environments, thus we give it the empty environment $\{\}$. Then we apply Definition 5.12, substituting the labeled version of addition. □

At dispatch, these new bounds are added to the labeled value sets $B_u$ and $B_l$, which maintain the non-dominated bounds.

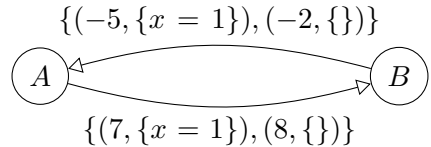

Figure 6.2: A dispatchable labeled distance graph fragment.

**Example 6.3** Consider the labeled distance graph fragment in Figure 6.2. If event $A$ is executed at $t = 2$, then we derive bounds $\{(7, \{x = 1\}), (4, \{\})\} \leq B \leq \{(9, \{x = 1\}), (10, \{\})\}$ □

Since the bounds may vary between choices, Drake cannot generally expect to obey all the possible constraints, but instead is only required to enforce all the constraints implied by at least one complete environment.

**Example 6.4** Assume an event has lower bounds represented by the labeled value set $A \geq \{(2, \{x = 1\}), (0, \{\})\}$. This implies that under any set of choices where $x = 1$, $A \geq 2$, and otherwise, $A \geq 0$ is sufficient. If the dispatcher executes $A$ at $t = 0$, then it may not select $x = 1$. Thus, if the dispatcher can restrict its choices to those that do not include $x = 1$, while leaving other consistent options, then it may execute $A$ at $t = 0$. If all the remaining consistent full assignments to the choices require that $x = 1$, then the dispatcher must wait until $t = 2$ before executing $A$. □

Drake performs this reasoning by collecting the environments for the bounds that a particular execution would violate, and by determining whether those environments can be made conflicts without making every complete environment inconsistent. If complete environments would remain consistent, the execution is performed and then environments are made conflicts; otherwise the execution is not possible, and discarded.

Finally, Drake simulates the collapse of zero-related events. At each time, Drake attempts to execute each event individually and also attempts to execute the sets of zero-related vertices recorded by the compiler. As noted before, zero-related sets may exist under some environments and not in others. If the zero-related set is enforced, then all of its member events must be executed together, or not at all. Therefore, if the dispatcher considers executing a set of events that is a strict subset of a particular zero-related set, leaving out at least one other member of the zero-related set, then it must discard the environments where the zero-related set is implied. Additionally, while there cannot be enablement constraints between zero-related events under any complete environments where the zero-related group holds, other environments may imply strict orderings between the events, which the dispatcher discards when executing the zero-group simultaneously. Hence, the dispatcher must discard any associated environments.

We summarize the rules as follows, which are further illustrated in Example 6.7.

**Theorem 6.5** *If $S$ is set with one event or a set of zero-related events in a Labeled STN, then the set may be executed at time $t$ if and only if there is at least one consistent complete environment where:*

1. *For every $A_l$ that is the lower bound labeled value set of some $A \in S$, such that $A_l \leq A$, for every $(l, e) \in A_l$ such that $l > t$, $e$ is a conflict.*

2. *For every $A_u$ that is the upper bound labeled value set of some $A \in S$, such that $A_u \geq A$, for every $(u, e) \in A_u$ such that $u < t$, $e$ is a conflict.*

3. *For every pair of events, $A \in S, B \notin S$, such that $B$ is not yet scheduled, for every $(w, e) \in W(A, B)$ such that $w < 0$, $e$ is a conflict.*

    *4. For every zero-related set of events $Z$ with environment $e$, if $S \subset Z$, $e$ is a conflict.*

    *5. For every $A_1, A_2 \in S$, for every $(w, e) \in W(A_1, A_2)$ such that $w < 0$, $e$ is a conflict.*☐

PROOF If there is a consistent full environment after creating these conflicts, then it is necessarily not subsumed by the environments of any of the constraints that imply the events cannot be executed at time $t$. Thus, the consistent full environment corresponds to a component STN where every constraint holds, so the events may be executed. If there is no consistent environment, then at least one of the above constraints prohibiting the proposed execution exists in each remaining component STN, so the execution is not valid. ☐

Similarly to the STN case, the dispatcher must check for missed upper bounds during every time step. In an STN, a missed upper bound implies the execution has failed. In a Labeled STN, a missed upper bounds implies that any complete environments subsumed by the environment of the missed upper bound are no longer valid.

**Theorem 6.6** *If event $A$ is not executed at time $t$, has upper bound labeled value set $A_u$, and $(u, e) \in A_u$ such that $u < t$, then $e$ is a conflict.* ☐

PROOF The theorem follows from noting that the upper bound exists in every component STN corresponding to a complete environment subsumed by $e$, thus every environment subsumed by $e$ is inconsistent, which by definition, makes $e$ a conflict. ☐

It is possible that missed upper bounds could invalidate all remaining complete environments, in which case dispatch has failed and the dispatcher should signal an error immediately.

**Example 6.7** Consider the dispatchable labeled distance graph in Figure 6.3, with events $A, B, C, D$ and choice variables $x, y \in \{1, 2\}$. There is a zero-related set $\{A, C\}$ under environment $\{x = 1\}$. Assume that the current time is $t = 0$ and no events have been executed yet. All the possible complete environments are initially consistent. Consider some possible executions and their consequences.

- Event $C$ has no predecessors, and no restrictions from its execution window. However, it is part of the zero-related set, so it may only be executed if we can make $\{x = 1\}$ a conflict, in order to remove the zero-related set from all possible executions. This is feasible, so we may execute $C$ at $t = 0$ and create the conflict. If we do this, then the ordering and inequality implied by edge $(A, D)$ is necessary for a consistent execution, as its environment cannot be made a conflict without making all complete environments inconsistent.

- Event $A$ has $D$ as a predecessor under $\{x = 2\}$, $C$ as a predecessor under $\{y = 1\}$, and is part of the zero-related set under $\{x = 1\}$. To execute $A$ at $t = 0$, we would need to make all three environments conflicts, but that would invalidate all possible choices, so we may not execute $A$ at $t = 0$.

- The zero-related set $\{A, C\}$ has no execution window restrictions from either $A$ or $C$, and has a predecessor $D$ under $\{x = 2\}$ from the edge $(A, D)$. Additionally, edge $(A, C)$ has a negative weight $-1$ with environment $\{y = 2\}$, so that environment is also made a conflict. Thus, we can create the conflicts $\{x = 2\}$ and $\{y = 2\}$ and execute both $A$ and $B$ at $t = 0$.

- Event $B$ has $A$ as a predecessor under $\{y = 1\}$ and $A$ is not yet executed, so we may make that environment a conflict and execute $B$. Assume that $A$ and $C$ are executed at $t = 0$, then the enablement constraint is satisfied, but any execution of $B$ before $t = 3$ requires making $\{y = 1\}$ a conflict. Additionally, if the current time grows to $t = 7$ and $B$ has not yet been executed, then the upper bound has been violated and $\{y = 1\}$ is then a conflict.

- Event $D$ has predecessor $C$ under environment $\{\}$, but $C$ has not yet been executed. Since making $\{\}$ a conflict makes all complete environments inconsistent, $D$ cannot be executed. □

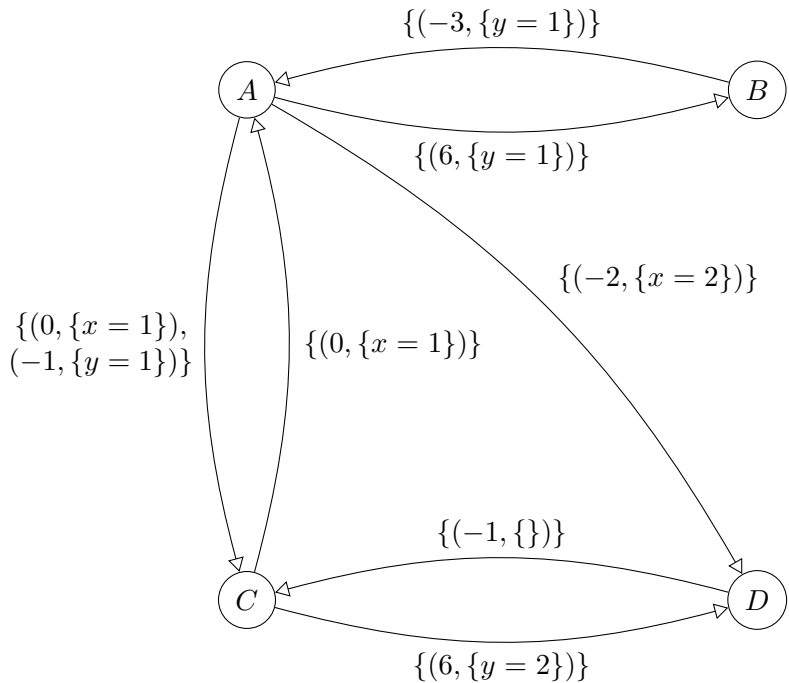

Figure 6.3: A dispatchable labeled distance graph.

This completes our presentation of Drake's dispatch algorithms. Essentially, it makes two adaptations to the STN dispatcher: (1) maintain labeled value sets for the execution windows of events and (2) allow the dispatcher to select between the choices by creating conflicts. The next section describes the compilation algorithm, which computes the dispatchable form of input Labeled STNs, ensuring that the local reasoning steps in the dispatching algorithm will satisfy the requirements of the plan.

## 7. Compiling Labeled Distance Graphs

We complete the description of Drake with the compiler, which reformulates the input Labeled STN into a form the dispatcher is guaranteed to execute correctly. The compiler leverages all of the labeling concepts we presented to efficiently compute a compact dispatchable form of input plans. An STN compiler takes a distance graph as its input and outputs another distance graph, the minimal dispatchable form of the input problem. Similarly, Drake's compiler takes a labeled distance graph as its input and outputs a labeled distance graph that is the minimal dispatchable form of the input.

Muscettola et al. (1998) introduced an initial compilation algorithm for STNs that operates in two steps. First, it computes the All-Pairs Shortest Path (APSP) graph associated with the STN, which is itself a dispatchable form. Second, the compiler prunes any edges that it can prove are redundant to other non-pruned edges, and which the dispatcher therefore does not need to make correct decisions. The pruned edges are *dominated*, and their removal significantly reduces the size of the dispatchable graph. The compiler tests for dominance by applying the following rule on every triangle in the APSP graph.

**Theorem 7.1 (Triangle Rule, Muscettola et al., 1998)** *Consider a consistent STN where the associated distance graph satisfies the triangle inequality; a directed graph satisfies the triangle inequality if every triple of vertices $(A, B, C)$ satisfies the inequality $W(A, B) + W(B, C) \geq W(A, C)$.*

*(1) A non-negative edge $(A, C)$ is upper-dominated by another non-negative edge $(B, C)$ if and only if $W(A, B) + W(B, C) = W(A, C)$.*

*(2) A negative edge $(A, C)$ is lower-dominated by another negative edge $(A, B)$ if and only if $W(A, B) + W(B, C) = W(A, C)$.* □

Although the basic concept of domination is unchanged, in that dominated constraints are not needed by the executive, the specifics are quite different here. Domination within a labeled value sets refers to labeled values that make other labeled values within that same set unnecessary. In the context of edge pruning, one edge could dominate another if the two edges share a start or end vertex, but not both. When we develop the labeled version of edge pruning, the labeled value from one edge weight labeled value set can dominate a value from a different edge weight labeled value set, where the corresponding edges share a start or end vertex.

**Example 7.2** Consider the two weighted graph fragments in Figure 7.1. In each case, edge $(A, C)$ is dominated, as shown by the theorem. □

The compiler searches for all dominated edges and then removes them all together. Some care is necessary to ensure that a pair of edges $AC$ and $BC$ do not provide justification for pruning each other; these edges are said to be *mutually dominant*. Typically, pruned graphs have a number of edges closer to $O(N \log N)$, than $N^2$, a significant savings. As a first prototype of Drake, Conrad, Shah, and Williams (2009) introduced a labeled extension of this algorithm. This extension was used to perform task execution on the ATHLETE Rover within the Mars Yard at NASA JPL.

Although simple and effective, this algorithm does not scale gracefully to large problems because it uses the entire APSP graph an intermediate representation, which is much larger than the final result. To resolve this, Tsamardinos, Muscettola, and Morris (1998) presented a fast compilation algorithm that interleaves the APSP step and the pruning step, and avoids ever storing the entire APSP graph. Their algorithm is derived from Johnson's algorithm for computing the APSP, which incrementally builds the APSP by repeated SSSP computations (Cormen, Leiserson, Rivest, & Stein, 2001). Johnson's algorithm uses Dijkstra's algorithm as an inner loop, providing it with faster performance than Floyd-Warshall for sparse graphs. Essentially, the fast compilation algorithm loops over the events in the graph, computes the SSSP for that event, and adds all non-dominated edges with the event as the source to the dispatchable form. Thus, this algorithm only needs to store the final dispatchable graph and one SSSP, avoiding the bloat in the intermediate representation. Thus, this algorithm has better space and time complexity than the APSP step followed by pruning.

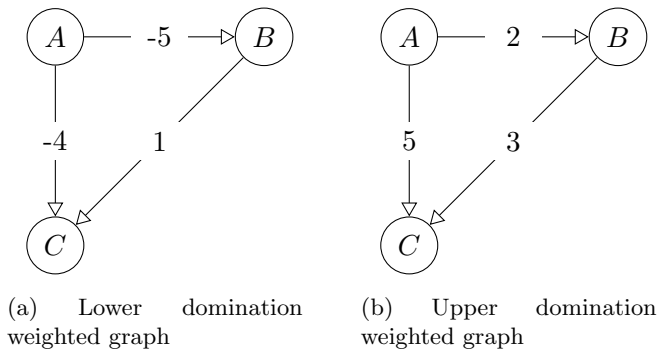

(a) Lower domination weighted graph

(b) Upper domination weighted graph

Figure 7.1: Examples of upper and lower dominated edges.

This paper introduces a novel extension to the Drake system that adapts the fast compilation algorithm to labeled graphs in order to avoid unnecessary storage growth during compilation. The fast algorithm requires a number of steps; this section describes each component of the original fast algorithm and its adaptation to the labeled setting. First, we recall Johnson's strategy to compute the APSP through iterated SSSP, and present the labeled adaptation of the SSSP algorithm. Second, we discuss the predecessor graphs that result from the SSSP and how to traverse them. Third, we describe how to interleave pruning with the repeated SSSP computations. Finally, we discuss the issues arising from mutual dominance, and the preprocessing step used to resolve them.

## 7.1 Johnson's Algorithm and the Structure of the Fast Algorithm

Many shortest path algorithms on weighted distance graphs essentially perform repeated relaxation over the graph. Floyd-Warshall loops repeatedly over the entire graph, relaxing the edges and computing the entire APSP in a single computation. Johnson's algorithm computes the APSP one source vertex at a time, using Dijkstra's SSSP algorithm as an inner loop to perform relaxations more efficiently. Since Dijkstra's algorithm only works for positively weighted graphs, Johnson's algorithm re-weights the graph before Dijkstra calls and un-weights it afterward. Unfortunately, adapting Dijkstra's algorithm to labeled distance graphs is inefficient because re-weighting both adds and subtracts labeled value sets, and both are not compatible with a single ordering. Therefore, we use Bellman-Ford's SSSP algorithm instead; we sacrifice the improved run-time of the fast algorithm, but preserve the lower space overhead. The fast STN compilation algorithm copies the essential structure of Johnson's algorithm. This section illustrates the overall algorithm in the unlabeled case with a partial example.

**Example 7.3** Consider one step of compilation shown in Figure 7.2. The input distance graph is shown in Figure 7.2a. When we compute the SSSP of the input for source vertex $B$, the output is the predecessor graph shown in 7.2b, which depicts the shortest distances and paths from $B$. The weights on the vertices indicate that the complete APSP graph should contain an edge from $B$ to $A$ with weight $-4$, $B$ to $C$ with weight $-1$, and $B$ to $D$ with weight 0. Furthermore, in the original graph, the shortest path from $B$ to $A$ uses the edge $B \rightarrow A$. The shortest path from $B$ to $C$ is either $B \rightarrow C$ or $B \rightarrow A \rightarrow C$; both have equal length. The shortest path from $B$ to $D$ is $B \rightarrow D$.

However, the dispatchable form of this problem does not actually need all three of these edges, which the compiler can deduce from the predecessor graph and SSSP values. Roughly, since $A$ has a lower distance from $B$ than $C$, and is along a shortest path to $C$, the edge $BC$ is not necessary. Therefore, the other two edges are inserted in the dispatchable graph, depicted in Figure 7.2c, and the edge $BC$ is discarded. This process is repeated for the other three vertices. □

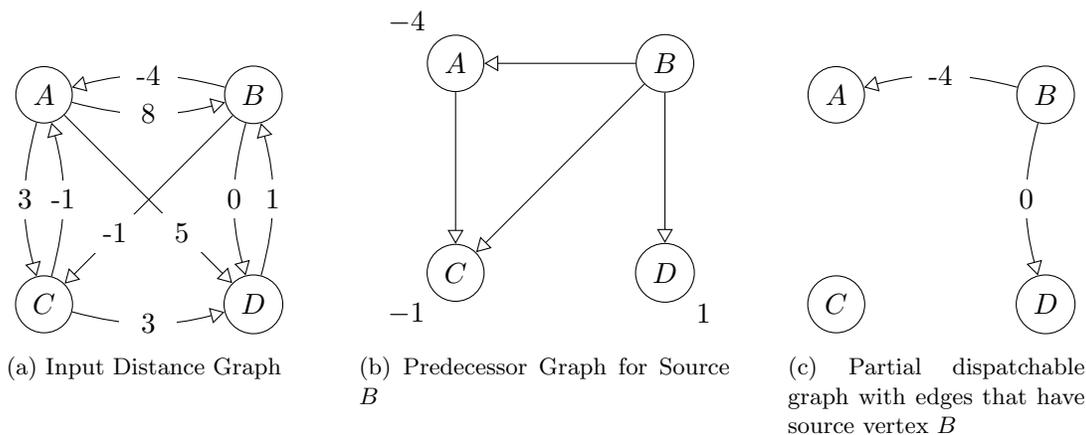

(a) Input Distance Graph

(b) Predecessor Graph for Source $B$

(c) Partial dispatchable graph with edges that have source vertex $B$

Figure 7.2: A single step of the fast reformulation algorithm for STNs

## 7.2 Labeled Bellman-Ford Algorithm

Drake uses the Bellman-Ford algorithm as a central building block for the variant of the fast STN algorithm Drake uses; Bellman-Ford derives the tightest possible edge weights, while simultaneously deriving the predecessor graph. This graph provides enough information to prune the dominated edges. To begin, we provide Algorithm 7.1, taken directly (with slightly altered notation) from the work of Cormen et al. (2001). The algorithm loops over the edges and performs relaxations, then tests for negative cycles to ensure the result is valid. The value $d[v]$ is the distance of the vertex $v$ from the input source vertex $s$. The value $\pi[v]$ is the vertex that is the (not necessarily unique) predecessor of $v$ when forming a shortest path from $s$ to $v$. Relating to Figure 7.2b, $d$ corresponds to the annotations next to the vertices, and $\pi$ specifies the directed edges.

---

**Algorithm 7.1** Bellman-Ford Algorithm

---

1: **procedure** BELLMANFORD($V, E, W, s$)
2:     INITIALIZESINGLESOURCE($V, W, s$)
3:     **for** $i \in \{1...|V|-1\}$ **do**                                   ▷ Loop for relaxation
4:         **for** each edge $(u, v) \in E$ **do**
5:             RELAX($u, v, W$)
6:         **end for**
7:     **end for**
8:     **for** each edge $(u, v) \in E$ **do**                           ▷ Check for negative cycles
9:         **if** $d[v] > d[u] + W(u, v)$ **then**
10:             **return false**                   ▷ Fail if a negative cycle is found
11:         **end if**
12:     **end for**
13:     **return true**
14: **end procedure**

15: **procedure** INITIALIZESINGLESOURCE($V, s$)
16:     **for** each vertex $v \in V$ **do**
17:         $d[v] \leftarrow \infty$
18:         $\pi[v] \leftarrow$ NIL
19:     **end for**
20:     $d[s] \leftarrow 0$
21: **end procedure**

22: **procedure** RELAX($u, v, W$)
23:     **if** $d[v] > d[u] + W(u, v)$ **then**
24:         $d[v] \leftarrow d[u] + W(u, v)$
25:         $\pi[v] \leftarrow u$
26:     **end if**
27: **end procedure**

---

As we will see in the next few sections, the fast compilation algorithm for STNs requires that we have access to the entire predecessor graph, which encodes all shortest paths in the graph, rather than any single path. To compute the entire graph, Drake makes each $\pi[v]$ a set, and if RELAX determines that $d[v] = d[u] + W(u, v)$, it then pushes $u$ onto $d[v]$. When the strict inequality holds and $d[v]$ is updated, $\pi[v]$ is set to the single element set $\{u\}$.

Now we discuss the changes necessary to adapt the Bellman-Ford algorithm to Labeled Distance Graphs. The relaxation procedure computes the new bound on the path length from $s$ to $v$, and if it dominates the old value, replaces the old value. If the vertex $u$ allows us to derive the current dominant value $d[v]$, then it is stored in $\pi[v]$. Drake's use of labeled value sets that store value pairs $(d, \pi) \in \mathbb{R} \times V$ allow it to perform both tasks at once, with one data structure.

To see how this works, consider an example:

**Example 7.4** In Figure 7.3a, edge $(A, B)$ forms a snippet of a labeled distance graph; consider computing the SSSP with source vertex $A$. We know that the initialization routine under every choice is to set $d[u] = 0$ for the source vertex and $d[u] = \infty$ for all others, and to set $\pi[u] = \text{NIL}$ for every vertex. In the combined notation, the initial value for the weight of the vertices is $(0, \text{NIL})$ or $(\infty, \text{NIL})$, labeled with the empty environment $\{\}$. These are the initial weights shown on the vertices.

The relaxation of this edge allows us to derive that $B$ is distance 8 from $A$ under environment $\{x = 1\}$, and has predecessor $A$. Hence, we add the labeled value pair $((8, A), \{x = 1\})$ to the labeled value set for the weight of $B$. This new value has a tighter value for $d$, and hence is kept, but does not dominate the old value, thus both are kept. This single insertion step updates both the distance of $B$ from $A$ and the predecessor of $B$ along the shortest path. The result is depicted in Figure 7.3b. $\qquad\square$

It should now be clear why we elected to keep non-identical values with the same numeric value: the labeled value sets can hold all the predecessors for the target vertices, and not just a single predecessor. Algorithm 7.2 shows the generalization of this example. Initialization is exactly as in the example, it assigns the initial values from the standard algorithm with empty environments. During relaxation, following Lemma 5.13, the potential new path lengths are the cross product addition of the weight of $u$ added to the weight of the edge from $u$ to $v$, labeled with the union of the environments, if the union is valid and not subsumed by any known conflicts. Naturally, when relaxing $u$ through an edge to $v$, $u$ is the predecessor. When adding this value to the labeled value set $d[v]$, the domination criterion is used to determine if this new path length is non-dominated, and to prune non-dominant path lengths and predecessors as appropriate.

The final modification of the original algorithm is in the test for negative cycles; a negative cycle is detected if another relaxation of any edge would further decrease the distance from the source to any other vertex. In the labeled case, some choices may have negative cycles and others may not. We begin the computation by computing the relaxation $d[u] + W(u, v)$ with the labeled addition operator, and discard the predecessor vertices from $d[u]$, thus producing a real valued labeled value set. If there is any environment where $d[v] > d[u] + W(u, v)$ holds, that environment is a conflict of the Labeled STN.

Finding the minimal conflicts in practice is somewhat convoluted. Consider each pair of labeled values $(d_v, e_v) \in d[v], (d_{uw}, e_{uw}) \in d[u] + W(u, v)$ in turn. If $d_v > d_{uw}$ then there might be a conflict, however, there may be some environments where a smaller, dominant value of $d[v]$ takes precedence over $d_v$, thus preventing the inequality from being satisfied. Hence, if it exists, the conflict we deduce is $e_v \cup e_{uw}$ after modifying the union to avoid any other environments $e'_v$ such that $(d'_v, e'_v) \in d[v]$ and $d'_v \leq d_{uw}$. It is possible that conflicts could invalidate all possible choices, which the executive tests for, before determining that the plan is dispatchable.

**Example 7.5** Consider the following values for the inequality test, given two binary choice variables:

$$d[v] = \{(2, \{y = 1\}), (4, \{\})\} \tag{14}$$

$$d[u] + W(u, v) = \{(3, \{x = 1\}), (5, \{\})\} \tag{15}$$

$$\tag{16}$$

We must find and make conflicts for any minimal environments that imply $d[v] > d[u] + W(u, v)$ holds. Each side of the inequality has two possible values, so we can consider four pairs of values that could satisfy the inequality. Three of the pairs do not satisfy the inequality, and thus can cause no conflict: $(2, \{y = 1\}) < (3, \{x = 1\})$, $(2, \{y = 1\}) < (5, \{\})$ and $(4, \{\}) < (5, \{\})$. The last pair, $(4, \{\}) > (3, \{x = 1\})$, satisfies the inequality, so this pair could imply that the union of corresponding environments, $\{\} \cup \{x = 1\}$, is a conflict. However, to reach $(4, \{\})$ for the left hand side, we skipped over a smaller value, $(2, \{y = 1\}) \in d[v]$. To correctly address this ordering, the conflict must also avoid the environment for the smaller value. Taking $\{\} \cup \{x = 1\}$ and avoiding $\{y = 1\}$ produces one environment, $\{x = 1, y = 2\}$, which is a valid environment, and hence it is a conflict. $\qquad \square$

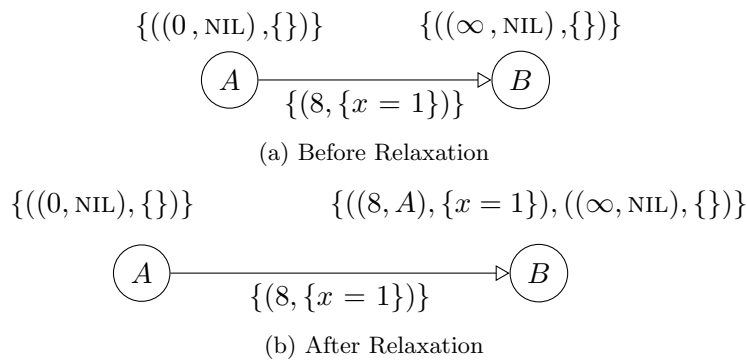

(a) Before Relaxation

(b) After Relaxation

Figure 7.3: A single labeled relaxation step.

## 7.3 Traversals of Labeled Predecessor Graphs

The fast compilation algorithm for STNs performs traversals of the predecessor graphs produced by the SSSP analysis, and checks each edge along the traversal for dominance. This is used to reduce the graph to minimal dispatchable form. In the unlabeled case, the $\pi[u]$ values specify a directed graph such as in Figure 7.2b, where we may use depth-first exploration to enumerate all possible paths beginning with the source vertex the SSSP is computed for. However, in the labeled case, more care is necessary.

The simplest way to understand which paths are valid under complete environments is to consider the projection of the predecessor graph under a particular complete environment $e$. Take every labeled value set $d[u]$ and query it under $e$. The result gives the shortest path distance and any predecessors, producing a standard unlabeled predecessor graph.

---

**Algorithm 7.2** Labeled Bellman-Ford Algorithm

---

1: **procedure** LABELEDBELLMANFORD($V, E, W, S, s$)
2:     LABELEDINITIALIZESINGLESOURCE($V, W, s$)
3:     **for** $i \in \{1...|V| - 1\}$ **do**             ▷ Loop over relaxations
4:         **for** each edge $(u, v) \in E$ **do**
5:             LABELEDRELAX($u, v, W$)
6:         **end for**
7:     **end for**
8:     **for** each edge $(u, v) \in E$ **do**           ▷ Test for negative cycles
9:         **for** each labeled value pair $(d_v, e_v) \in d[v]$ **do**
10:             **for** each labeled value pair $(d_{uw}, e_{uw}) \in d[u] + W(u, v)$ **do**
11:                 **if** $d_v > d_{uw}$ **then**
12:                     ADDCONFLICT($e_v \cup e_{uw}$, split to avoid all $e'_v$ such that
13:                     $(d'_v, e'_v) \in d[v]$ and $d'_v \leq d_{uw}$)
14:                 **end if**
15:             **end for**
16:         **end for**
17:     **end for**
18:     ISSOMECOMPLETEENVIRONMENTCONSISTENT?()
19: **end procedure**

20: **procedure** LABELEDINITIALIZESINGLESOURCE($V, s$)
21:     **for** each vertex $v \in V \setminus s$ **do**
22:         $d[v] \leftarrow \{((\infty, \text{NIL}), \{\})\}$
23:     **end for**
24:     $d[s] \leftarrow \{((0, \text{NIL}), \{\})\}$
25: **end procedure**

26: **procedure** LABELEDRELAX($u, v, W$)
27:     **for** labeled value pair $((d_u, \pi_u), e_u) \in d[u]$ **do**
28:         **for** labeled value pair $((d_w, \pi_w), e_w) \in W(u, v)$ **do**
29:             ADDTOLVS($d[v], ((d_u + d_w, u), e_u \cup e_w)$)
30:         **end for**
31:     **end for**
32: **end procedure**

---

**Example 7.6** Consider the labeled graph fragment in Figure 7.4a, where the vertices are labeled with their SSSP weights for source vertex $A$. Further consider the predecessor graph and path lengths implied under environment $e = \{x = 1, y = 1\}$: $d[B](e) = (2, A)$ and $d[C](e) = (1, A)$. Hence, both vertices $B$ and $C$ have $A$ as their only predecessor, and $B$ is distance 2 from $A$, and $C$ is distance 1 from $A$, as shown in Figure 7.4b. Vertex $D$ is distance 6 from $A$, with predecessor $C$. An important feature of this example is that for this environment, $C$ is not a predecessor of $B$, even though the pair $(7, C)$ is labeled with $\{y = 1\}$, which subsumes $e$. Predecessors are only provided by the dominant path length.□

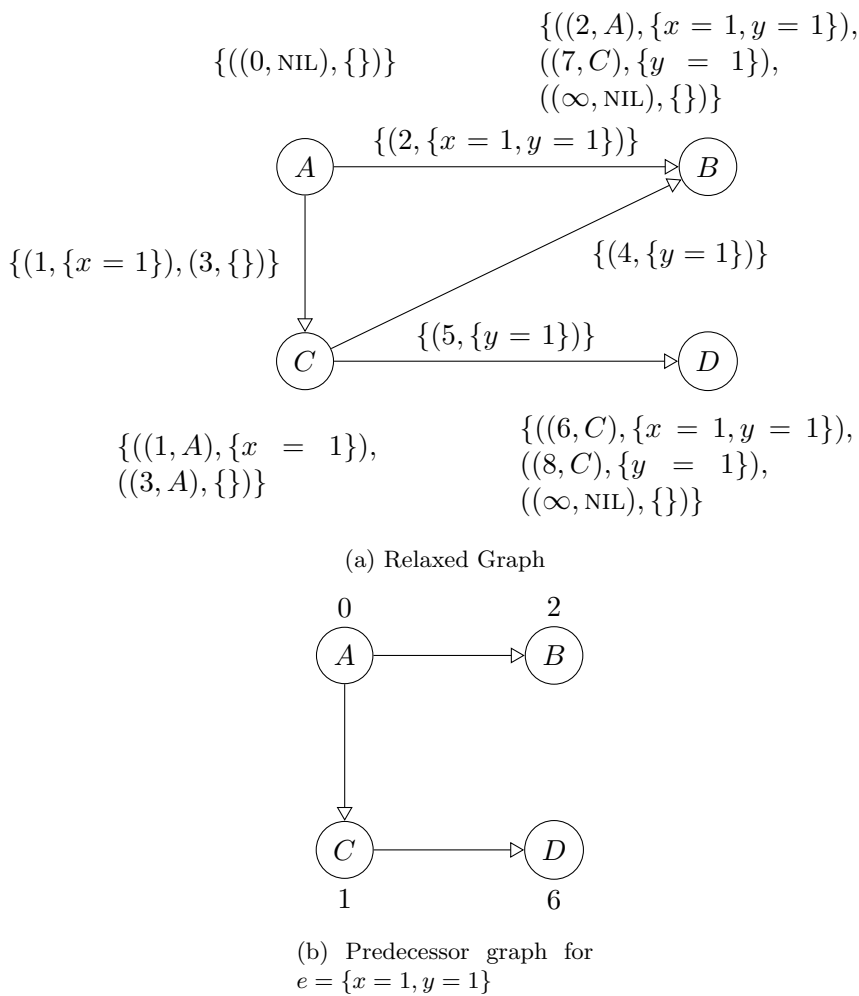

(a) Relaxed Graph

(b) Predecessor graph for $e = \{x = 1, y = 1\}$

Figure 7.4: Extracting predecessor graphs from labeled SSSP

Although it is instructive to consider the projection of the SSSP onto an environment for intuitive understanding, an implementation based on this approach is not efficient in space or time. As in the rest of this work, Drake directly performs the traversals on the labeled representation. Since paths in the predecessor graphs may exist under some environments and not others, we take the natural step and use minimal environments to specify when a particular path exists. Intuitively, we expect that the environment of a path is simply the union of the environments of the labeled value pairs that specified the predecessor edges used. This is true, but we require an extra step to test the validity of the path for that environment.

Consider a partial path from the source vertex $S$, which passes through some subset of vertices $X_1 \rightarrow \dots X_n$, constructed from labeled value pairs $((d_i, X_i), e_i) \in d[X_{i+1}]$, which represent some path through the predecessor graph. Then the path is valid if, for every vertex along the path, $(d_i, X_i) \in d[X_{i+1}](e_1 \cup \dots e_n)$, and the path environment is $e = e_1 \cup \dots e_n$. Essentially, this tests that labeled value pairs used to construct the path are actually still entailed by the union of all of their respective environments. If a path is invalid, all possible extensions are also invalid, because the same inconsistency will exist in all extensions.

**Example 7.7** We can apply this criteria to determine all the paths given by the SSSP in Figure 7.4a. Begin with the source vertex, $A$, which clearly exists under all environments, so we assign this partial path the empty environment. $B$ contains the labeled value pair $((2, A), \{x = 1, y = 1\})$, and thus is a candidate to extend our path. The union of the two environments is $\{\} \cup \{x = 1, y = 1\} = \{x = 1, y = 1\}$. When queried with this environment, $d[B]$ returns the labeled value pair that we used to propose the extension, so the path $A \rightarrow B$ has environment $\{x = 1, y = 1\}$. $B$ has no other possible extensions.

Returning to $A$, it has a possible extension to $C$ because of the labeled value pairs $((1, A), \{x = 1\})$ and $((3, A), \{\})$. First, consider $((1, A), \{x = 1\})$, and the query of $d[C]$ under $\{x = 1\}$ shows that the path exists, so we have the partial path $A \rightarrow C$ with environment $\{x = 1\}$. This path has a potential extension to $B$ because of the labeled value pair $((8, C), \{y = 1\})$. However, the query $d[B](\{x = 1\} \cup \{y = 1\}) = (2, A) \neq (7, C)$, so the extension is not valid, as expected from our earlier discussion. We may instead extend this path to $D$, giving the path $A \rightarrow C \rightarrow D$ the environment $\{x = 1, y = 1\}$. The queries on both $d[C]$ and $d[D]$ give the values used to generate the path, and hence the path is valid.

Returning to $A$, we consider the other path to $B$, using the labeled value pair $((3, A), \{\})$. Now the extension to $B$ passes the query test, so $A \rightarrow C \rightarrow B$ provides path length 7 if $\{x = 1\}$ but not $\{y = 1\}$. We label this path with the label $\{x = 1\}$, but must take care in later algorithms not to use it to imply that 7 is the shortest path length for $\{x = 1, y = 1\}$. Finally, we can extend the path to $D$, giving path $A \rightarrow B \rightarrow D$ with length 8 under $\{y = 1\}$, with the same caveat. □

Given this rule specifying the correct extensions to partial paths, we can construct a depth-first search to enumerate all possible paths in the acyclic predecessor graph, re-testing at each step to ensure the path is valid.

## 7.4 Pruning Dominated Edges with the SSSP

With the strategy for finding labeled paths through the labeled predecessor graphs, the extension of the pruning algorithm from the fast STN compilation algorithm is straightforward. Tsamardinos, et al. (1998) provide two theorems relating dominance of edges to paths in the predecessor graph, adjusted slightly for our notation. In the following, $A$ is assumed to be the source vertex of the SSSP, and $B$ and $C$ are other vertices.

**Theorem 7.8** *A negative edge $(A, C)$ is lower-dominated by a negative edge $(A, B)$ if and only if there is a path from $B$ to $C$ in the predecessor graph for $A$.* □

**Theorem 7.9** *A non-negative edge $(A, C)$ is upper-dominated if and only if there is a vertex $B$, distinct from $A$ and $C$, such that $d[B] \leq d[C]$ and there is a path from $B$ to $C$ in the predecessor graph for $A$.* □

**Example 7.10** Figure 7.5 shows two simple examples on which we can apply these two theorems. First, Figure 7.5a is a weighted distance graph and Figure 7.5c shows its predecessor graph for source vertex $A$, on which we apply Theorem 7.8. Edge $(A, C)$ is lower dominated because the weight $-5$ on $B$ implies an edge $(A, B)$ with weight $-5$, which is negative as the theorem requires. Furthermore, the predecessor graph has a path from $B$ to $C$ in the predecessor graph for $A$. Therefore, edge $(A, C)$ is dominated and needed in the dispatchable form.

Figures 7.5a and 7.5c similarly exhibit Theorem 7.9. There is a path $A \to B \to C$ in the graph, and $d[B] = 2 \leq d[C] = 5$, so edge $(A, C)$ is upper-dominated. Thus, edge $(A, C)$ is not needed in the dispatchable form.

In both these examples, this step derives every possible edge weight for edges in the dispatchable form with $A$ as their source, namely edges $(A, B)$ and $(A, C)$, and determines that only $(A, B)$ is actually needed. □

In the labeled case, particular labeled edge weights are dominated if the above conditions hold under all the environments the weight holds in.

**Theorem 7.11** *A negative labeled edge weight $(d_C, e_C)$ in $d[C]$ is lower-dominated by a negative labeled edge weight $(d_B, e_B)$ in $d[B]$ if and only if $e_B = e_C$ and there is a path from $B$ to $C$ with environment $e_P$ in the predecessor graph for $A$ such that $e_P = e_C$.* □

**Theorem 7.12** *A non-negative labeled edge weight $(d_C, e_C)$ in edge $d[C]$ is upper-dominated if and only if there is a labeled edge weight $(d_B, e_B)$ in $d[B]$, such that $B$ is distinct from $A$ and $C$, $e_B$ subsumes $e_C$, $d_B \leq d$, $C$, and there is a path from $B$ to $C$ with environment $e_P$ in the predecessor graph for $A$ such that $e_P = e_C$.* □

In practice, Drake searches over every path in the labeled predecessor graph with the source vertex as its start, and applies these theorems to find dominated edges. Specifically, during the traversal, it records the smallest vertex weight of any vertices along the path, not counting the source. That value is compared to other vertex weights of extensions of the path to apply the domination theorems. Every time a vertex weight is found to be dominated with some path, it is recorded in a list. After all traversals are done, every labeled value in the vertex weights not present in the list of dominated values is converted into an edge in the output dispatchable graph.

These two theorems require that $e_P = e_C$ because the path must hold under all environments where the value $d_C$ does, but we also do not want $e_P$ to be tighter. Recall that the vertex weights we might prune also specify the paths. If $e_C$ is tighter than $e_P$, then it must have a lower path length than the one implied by $e_P$, or else it would not be in the labeled value set $d[C]$. Thus, we cannot guarantee that our path is the shortest path from the source to $C$, so this path is not suitable to prune it.

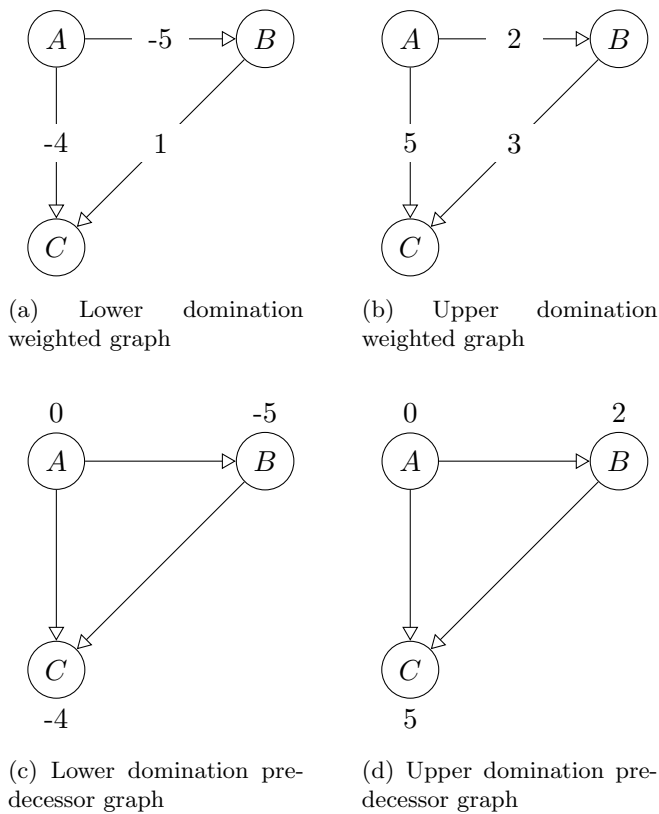

(a) Lower domination weighted graph

(b) Upper domination weighted graph

(c) Lower domination predecessor graph

(d) Upper domination predecessor graph

Figure 7.5: Simple edge domination examples.

**Example 7.13** To demonstrate the application of these ideas, reconsider Figure 7.4a. Example 7.6 gave the possible paths in the graph. The first path is $A \to B$ with path environment $\{x = 1, y = 1\}$. After reaching $B$, the minimal vertex weight is 2, but there are no extension of this path, so nothing can be pruned. In general, the first step from the source vertex cannot be pruned.

The next step of the traversal reaches $C$ with environment $\{x = 1\}$, and the minimal path length is 1. $C$ cannot be pruned. This path cannot be extended to $B$, but it does have an extension to $D$, using the vertex weight $((6, C), \{x = 1, y = 1\})$. This path length is strictly longer than the path length of 1, and the environment of the value is equal to the environment of the path, so we add it to the pruned list.

Next consider path $A \to C$ under environment $\{\}$ and path length 3. The extension to $B$ using $((7, C), \{y = 1\})$ is also prunable. Note that this path could never be used to prune the shorter path length $((2, A), \{x = 1, y = 1\})$ because $\{x = 1, y = 1\} \neq \{y = 1\}$. Likewise, the extension to $D$ prunes $((8, C), \{y = 1\})$.

Collecting the non-pruned edges means that the algorithm adds edge $(A, B)$ to the output dispatchable graph with weight $W(A, B) = \{(2, \{x = 1, y = 1\})\}$, and adds edge $(A, C)$ with weight $W(A, C) = \{(1, \{x = 1\}), (3, \{\})\}$. We drop infinite weights, allowing them to be implicitly specified, and both finite weights of $D$ were pruned, so it does not add the edge $(A, D)$ at all. □

Essentially, the pruning algorithm has the same structure as the unlabeled fast compilation algorithm. The major difference is that the values to prune have environments and that paths in the predecessor graph only exist under particular environments. Thus, the pruning step must satisfy the pruning requirement with the identical environment to prune a labeled value.

## 7.5 Mutual Dominance and Rigid Components

Tsamardinos et al. (1998) showed that rigid components in the distance graph of an STN create mutual dominance, where two edges $(A, B)$ and $(A, C)$ are each used as evidence to prune the other, and both are incorrectly removed when the algorithm of the previous section is applied. The correct solution is to perform either pruning, but not both. Unfortunately, it is difficult to test for mutual dominance during the SSSP pruning step. Instead, they present a pre-processing step that identifies the rigid components, updates the output dispatchable graph accordingly, and alters the input weighted distance graph to remove the rigid components entirely, before beginning the process of computing edges in the dispatchable form through repeated SSSPs. This leaves a problem with no mutually dominated edges, and thus the pruning step can prune all edges found to be dominated without any risk of encountering this problem. This section adapts Tsamardinos et al.'s approach by identifying labeled rigid components and uses a labeled analogue of the alteration process.

**Example 7.14** Consider the predecessor graph fragment in Figure 7.6. There is a path $A \rightarrow B \rightarrow C$, with $d[B] = 1$ and $d[C] = 1$, so that $d[B] \leq d[C]$. Thus, the edge $(A, C)$ can be pruned. Likewise, there is a path $A \rightarrow C \rightarrow B$ that allows us to prune $(A, B)$ from the dispatchable graph. Then the result is that $A$ is not constrained to the rest of the events, which would allow the dispatcher to select an inconsistent schedule. These edges are mutually dominant. Our algorithm is able to prune both edges because there is a path $B \rightarrow C$ and $C \rightarrow B$, if either did not exist, only one pruning would take place. □

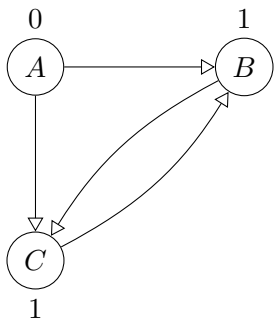

Figure 7.6: A predecessor graph, $(A, B)$ and $(A, C)$ are mutually dominant.

Tsamardinos et al. (1998) show that mutual dominance occurs if and only if there are rigid components in the problem, which are collections of events that must be scheduled with a fixed difference in their execution times. In Figure 7.6, $B$ and $C$ are constrained to occur at the same time, so they are rigidly connected. Rigidly connected components in a consistent distance graph coincide with strongly-connected components in the predecessor graph for a single source shortest path for arbitrary connected vertices. Thus, they show that we can identify all instances of possible mutual dominance by finding all the strongly-connected components of the predecessor graphs, and then remove them.

Intuitively, we might understand the issue as the input problem having less degrees of freedom than it appears to have. If two events are actually constrained to occur at the same instant, then the dispatcher has one decision, not two. When compiling, each decision looks redundant to the other, leading to incorrect prunings, as in Example 7.14. The solution is to remove the redundancy by replacing the entire rigid component with a single vertex, called the *leader*. The prior work actually removes all events other than the leader with a preprocessing step, contracting the rigid component to a single vertex. Since we are working with labeled distance graphs, where rigid components may exist only in some environments, we must replicate this effect indirectly, in two steps: provide information in the compiled form so the dispatcher can respect the rigid component, and alter the input graph so that under conditions where the rigid component exists, the compilation algorithm only reasons over the leader.

The first step of preprocessing is to identify the rigid components. Arbitrary graphs might not be connected, so an SSSP from any arbitrary vertex may not reach all the vertices, and thus cannot find rigid components in the disconnected subgraph. A standard strategy, as in the Johnson's algorithm preprocessing step, is to introduce an extra vertex for the preprocessing step, connected to every vertex with a zero weight edge. This extra vertex is connected to every vertex, which makes it an ideal source to search from, even for disconnected graphs. Furthermore, the extra vertex does not alter the rigid components in the predecessor graph. In the labeled case, the equivalent approach is to introduce a new vertex connected to every other vertex with an edge with weight $(0, \{\})$, so that the zero weight edge applies under every environment. Then, the SSSP is computed with the added vertex as the source. Finally, the algorithm searches the predecessor graph for cycles. The simplest, if not necessarily the most efficient, method in the labeled case is simply to use our knowledge about finding paths in the predecessor graph to search for loops.

As we might expect, rigid components in Labeled STNs are given labels, since they may only exist under some environments. We need to find the maximal rigid components with minimal environments. Thus, if there is a rigid component $\{A, B, C\}$, we do not separately identify $\{A, B\}$ as a rigid component. For every vertex in $V$, run a depth first search for a path back to the vertex. If found, the vertices on the path are a rigid component, and the path environment is the environment of the rigid component. Note that a valid path may visit any vertex up to twice, as in the example below. As the search finds rigid components, it maintains a list of the maximally sized rigid components with minimal environments.

**Example 7.15** Figure 7.7 shows a small labeled distance graph with vertices $\{A, B, C\}$. Recall that having an opposing pair of edges, one with the negative weight of the other, implies a rigid component, so this problem obviously has some.

To apply our technique, the algorithm first adds another vertex $X$, and connects it to each vertex with an edge with weight $(0, \{\})$. Figure 7.7b also shows the predecessor graph computed after this has been done. Second, the algorithm searches for loops in the graph. It begins with $A$: the path $A \rightarrow B \rightarrow A$ with environment $\{x = 1\}$ is a loop, as is $A \rightarrow B \rightarrow C \rightarrow B \rightarrow A$ with environment $\{x = 1, y = 1\}$. These paths imply rigid components $\{A, B\}$ with environment $\{x = 1\}$ and $\{A, B, C\}$ with environment $\{x = 1, y = 1\}$, respectively. Note that the second path visits $B$ twice. From $B$, it find paths $B \to A \to B$ with environment $\{x = 1\}$, which is equivalent to the one found from $A$, and $B \to C \to B$ with environment $\{y = 1\}$, which is a new rigid component. The paths from $C$ re-derive the same rigid components. Thus, there are three maximal rigid components, $\{A, B\}$ with environment $\{x = 1\}$, $\{B, C\}$ with environment $\{y = 1\}$, and $\{A, B, C\}$ with environment $\{x = 1, y = 1\}$.

If the two edges between $B$ and $C$ had the environment $\{x = 1\}$ instead of $\{y = 1\}$, then there would only be a single maximal rigid component $\{A, B, C\}$ with environment $\{x = 1\}$. The smaller components, such as $\{A, B\}$ with $\{x = 1\}$, would be non-maximal and not needed. □

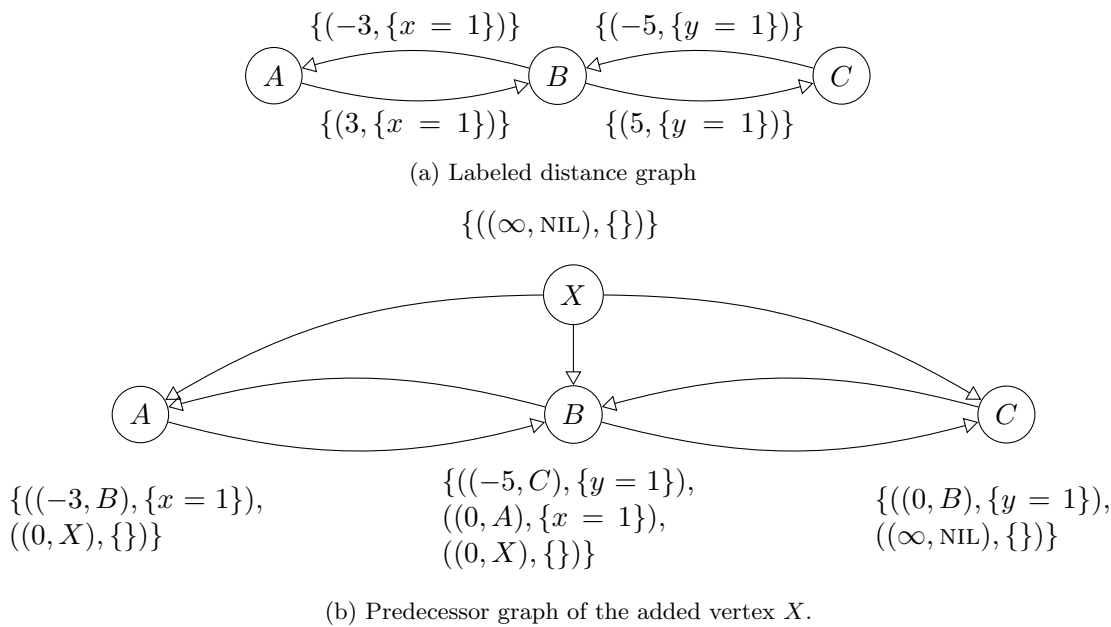

(a) Labeled distance graph

(b) Predecessor graph of the added vertex $X$.

Figure 7.7: Identifying rigid components in labeled distance graphs.

After completing the search for rigid components, we must process them. The first processing step is to identify a leader, the first vertex to occur, which we will use to represent the entire rigid component. The leader is the vertex with the lowest distance from the added vertex, queried under the environment of the rigid component. Ties may be broken arbitrarily.

The second processing step is to update the dispatchable form accordingly. Within the rigid component, the events are rigidly bound together and the leader is executed before any other event of the rigid component. Thus, in the dispatchable form, we constrain the non-leader events to occur the correct, fixed amount of time after the leader. Thus, if $A$ is the leader of the rigid component, and $B$ is some other vertex of the rigid component with environment $e$, the the output graph should have the labeled edge weight $(d[B](e) - d[A](e), e)$ on edge $(A, B)$ and $(d[A](e) - d[B](e), e)$ on edge $(B, A)$. If any events are rigidly connected in a way such that they will be executed at the same time, the dispatcher needs them listed explicitly, because the execution window and enablement tests alone do not guarantee correct execution of such plans. During this step we may identify maximal groups of events that are constrained to occur at the same time as the leader, or the same fixed duration after the leader. For example, if $A$ is the leader and both $B$ and $C$ follow $A$ by exactly three time units, then $B$ and $C$ are constrained to occur at the same time. As usual, these zero-related vertices are recorded with a label, which is the same environment as the rigid component.

Third, we need to alter the labeled distance graph so that the rigid component no longer appears to exist, but is instead totally represented by the leader vertex. The algorithm begins with edges that are interior to the rigid component. If $(A, B)$ are both vertices in a rigid component with environment $e$, then the algorithm prunes $W(A, B)$ and $W(B, A)$ of any environments subsumed by $e$. If $(d, e_d) \in W(A, B)$, then if $e$ subsumes $e_d$, the algorithm removes the value pair. If $e$ does not subsume $e_d$, then replace $(d, e_d)$ with labeled values $(d, e'_d)$ where $e'_d$ avoids $e$, which may require multiple new values. Repeat for $W(B, A)$.

Finally, we need to adjust edges that enter or leave the rigid component to ensure the input to the compiler has no cycles in any predecessor graphs. The idea is to move these edges to the leader under the environments of the rigid component, and to remove them from the non-leader. Assume $A$ is the leader of a rigid component with environment $e$, $B$ is another vertex of the rigid component, and $C$ is a vertex not in the rigid component. For every labeled value $(d, e_d) \in w(B, C)$ for some vertex, the algorithm puts $(d + d[B](e) - d[C](e), e_d \cup e)$ in $W(A, C)$. Next, it replaces $(d, e_d) \in W(B, C)$ with values $(d, e'_d)$ for each possible $e'_d$ that is a union of $e_d$ and the constituent kernels of $e$. The algorithm repeats for $W(C, B)$, putting $(d + d[C](e) - d[B](e), e_d \cup e)$ in $W(C, A)$, and replacing $(d, e_d)$ with $(d', e'_d)$ avoiding $e$. Unfortunately, this strategy leads to duplication that grows linearly with the domain size of the choice variables. In practice, though, many problems do not have any rigid components, or they are limited and the penalty is minor. Our results show some outliers that may be associated with this growth, but most problems are not affected.

**Example 7.16** As usual, let $x, y \in \{1, 2\}$ be choice variables. Figure 7.8a depicts an input labeled distance graph fragment. From the opposing edge weights 1 and $-1$, we can easily identify that $\{A, B\}$ is a rigid component under environment $\{x = 1\}$. We need to follow the above procedure to process it.

First, $A$ is always scheduled before $B$, and is therefore the leader. If we ran the SSSP on an added vertex, we would find that $d[A](\{x = 1\}) = -1$ and $d[B](\{x = 1\}) = 0$, proving this assertion. Since $d[B](\{x = 1\}) - d[A](\{x = 1\}) = 1$, event $B$ follows $A$ by one time unit. Therefore, edges are inserted into the output dispatchable form shown in Figure 7.8c that enforce this delay.

Next, we alter the input graph to remove the rigid component, beginning with edges between the rigid component's vertices. Edge $(B, A)$ has one weight on it, which is subsumed by the environment of the rigid component, so it only exists when the rigid component does, and thus has already been handled, and is pruned. Likewise, the weight $(1, \{\}) \in W(A, B)$ is pruned. On the other hand, the weight $(2, \{x = 2\}) \in W(A, B)$ already avoids the environment of the rigid component, so is left unchanged.

Continue with edges that enter or leave the rigid component. $(A, C)$ uses the leader vertex, so does not directly require any modification. Edge $(B, C)$ needs to be moved, giving a new weight $(4 + 1, \{x = 1\})$ on $W(A, C)$, which is already present, and requires no modification. Since its environment is subsumed by the rigid component environment, $(4, \{x = 1\})$ is removed from $w(B, C)$. Edge $(C, B)$ also needs to be moved. It leads to a new value $(2 - 1, \{y = 1\} \cup \{x = 1\}) = (1, \{x = 1, y = 1\}) \in W(C, A)$. Since $\{y = 1\}$ is not subsumed by the rigid component's environment, we must modify the value on $(C, B)$ so that it avoids the environment. Avoiding $\{x = 1\}$ means that $\{x = 2\}$, so we modify the value $(2, \{y = 1\}) \rightarrow (2, \{x = 2, y = 1\})$. These modifications completely remove the rigid component and provide the dispatcher with sufficient information to correctly execute the rigid components we have removed. □

After these modifications the labeled distance graph is guaranteed not to have any cycles in the predecessor graphs, and therefore has no rigid components. Thus the rest of the fast algorithm given in this section correctly compiles it to dispatchable form. We adjusted the unlabeled algorithm to search for labeled rigid components. Additionally, the unlabeled algorithm removes the non-leader vertices of rigid components, moving and modifying edges as necessary enforce the original input constraints. Since those vertices may be needed under environments where the rigid component does not exist, Drake instead modifies the process of moving edges to replicate the effect of removing the non-leader vertices.

Drake's compilation algorithm is designed to use labeling concepts to compute a compact version of the dispatchable form of plans with choice. The structure of this algorithm is similar to the unlabeled version, but a number of modifications are made within each step to reason about environments. This section completes the presentation of Drake's algorithms.

## 8. Results

Finally, we explore Drake's performance, both from a theoretical standpoint and experimentally. Our analysis gives some justification for why we expect Drake's representation to be compact, and our experimental results give evidence that Drake performs as intended.

### 8.1 Theoretical Results

We give a brief characterization of the analytical worst case performance of Drake's algorithms. The direct enumeration of STNs, as in Tsamardinos et al.'s (2001) work, uses one STN for each consistent STN. If there are $n$ choices with $d$ options each, and $v$ vertices, and we assume that the compiled sparse graphs are of size $\mathrm{O}(v \log v)$, then the compiled size is $\mathrm{O}(n^d v \log v)$. In contrast, Drake does not store the component STNs independently, but only stores the distinct values, so the size is $\mathrm{O}(kv \log v)$, where $k \leq n^d$. In the worst case, where every single component STN is completely different, there are no similarities between choices to exploit, hence Drake's compiled representation is the same size, but is never worse, up to a constant.

There is a strong parallel to the existing theory about the tree width of general constraint satisfaction problems. Dechter and Mateescu (2007) explain that while a general constraint satisfaction problem with $n$ variables of domain size $d$ can in general only be solved in $\mathrm{O}(n^d)$ steps, many problems have more structure. The *tree width*, $n^* \leq n$ of the problem represents the number of variables that effectively interact, so that the search can be completed in $\mathrm{O}((n^*)^d)$ time. Similarly, in Labeled STNs, the choice variables may not interact fully, and thus there is an effective number of choice variables in the problem that is often smaller than the total number. In this notation, Drake's compiled problems have size $\mathrm{O}\big((n^*)^d n \log v\big)$. The smaller base of the exponent can lead to significant savings.

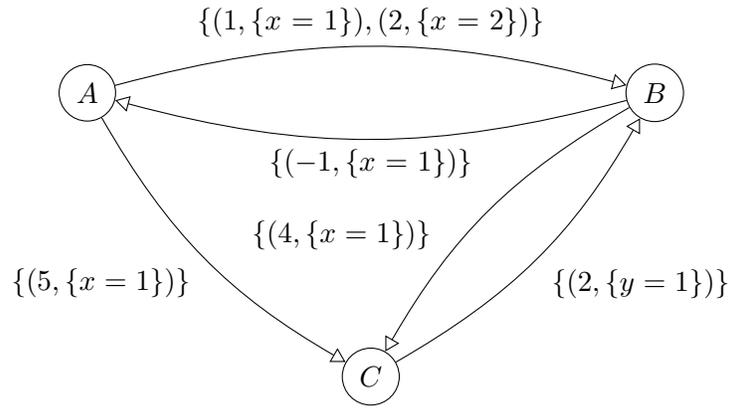

(a) Input labeled distance graph

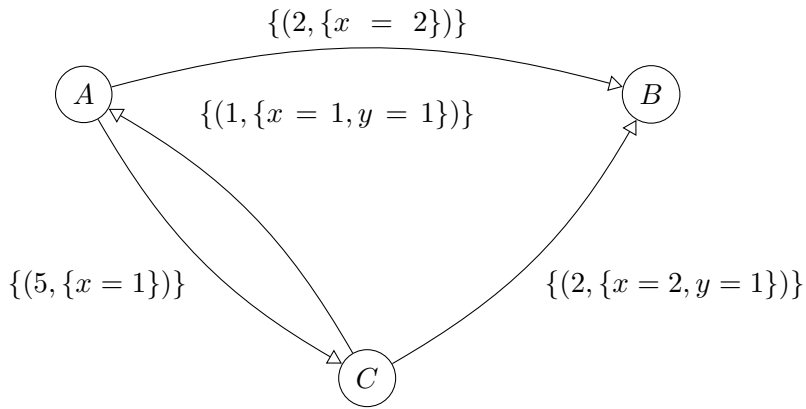

(b) Contracted labeled distance graph

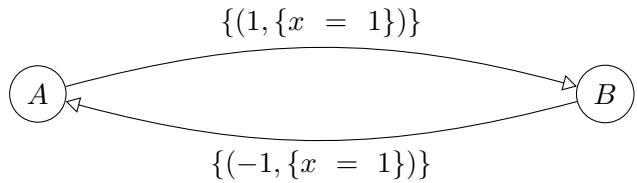

(c) Dispatchable form of rigid component

Figure 7.8: An example of processing a rigid component from a Labeled Distance Graph.

The compile time and run-time latency are more difficult to characterize, however, because the overhead of the labeling operations also grows with $n*$. Therefore, we do not attempt an analytic analysis.

## 8.2 Experimental Results

This section presents an experimental validation of Drake's compilation and dispatch algorithms on randomly generated, structured problems. First, we develop a suite of random structured Labeled STNs, derived from Stedl's (2004) problem generator. Then we compile and dispatch the suites of problems twice, once with Drake and once by explicitly enumerating all component STNs, following techniques developed by Tsamardinos et al. (2001). Finally, we compare the compiled size of the problems, the compilation time, and the execution latency. Throughout this section, our plots use the number of consistent component STNs as the horizontal axis, because it appears to correlate well with the effective difficulty of the problem. For each of the metrics, we provide the results from the two methods side by side in one plot, and show the ratio of performance in another one, to allow point-wise comparison of the difference in performance between identical problems. The problems are constructed from 2-11 binary choices or 2-7 ternary choices. There are 100 problems at each of these sizes; the problems range from 4 consistent component STNs up to about 2000. The number of events ranges from 4 to about 22 as the number of component STNs increase to keep a consistent ratio of constraints to events.

This comparison is performed with a Lisp implementation, run on a 2.66 GHz machine with 4 Gb of memory. There are some performance related implementation details we have omitted in prior sections. For example, labeled value sets are stored as ordered lists to reduce insertion and query time. Additionally, we found that memoization of subsumption and union operations dramatically improved performance. The implementation aggressively prunes values with inconsistent environments to avoid any unnecessary reasoning. The STN compiler and dispatcher exercise the same code as Drake to support a fair comparison, and pays a small overhead in execution speed.

The first metric of comparison is the size of the dispatchable form of the random problems. We computed this by serializing the graph representations to strings. Since Drake's compilation algorithm is derived from the fast STN compiler, the maximum memory footprint for both compilation and dispatch is no more than about double these numbers.

Since we designed Drake with this metric in mind, we expect the improvement to be clear and significant, and this is what Figure 8.1 shows. In the ratio plot, the value is the multiple of improvement of Drake over STN enumeration. Except for small problems, Drake's memory performance is superior, and the improvement ranges up to around a 700 times smaller memory footprint for the largest problems. The STN enumeration Tsamardinos et al. (2001) develops uses up to around 2 MB of storage, and although that is insignificant for a modern desktop, it is often significant for embedded hardware, especially if the system must store a library of compiled plans. In contrast, Drake's memory footprint is 1-10 kB for most cases, which is trivial, even in large numbers, for any hardware. More than half the worse performing examples, those that do not fit in the main band of results, correspond to ternary choices. We believe this likely corresponds to the growth caused by avoiding environments when handling complex or overlapping rigid components in these problems.

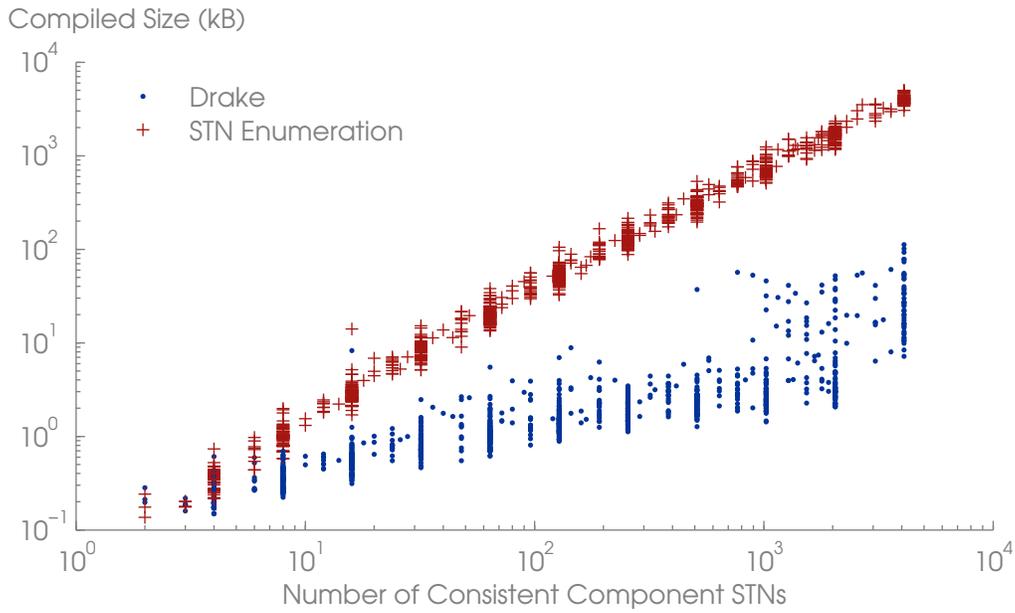

(a) The size of the dispatchable labeled distance graphs.

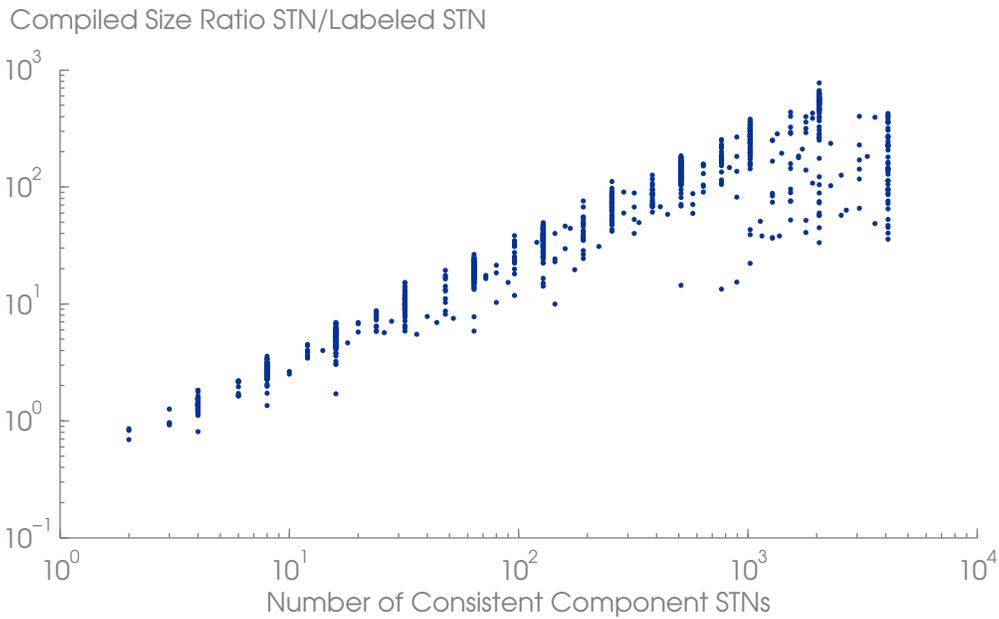

(b) The ratio of the size of the compiled STN enumeration size to the size of Drake's labeled distance graph.

Figure 8.1: The size of the dispatchable form of the random problems as a function of the number of component STNs.

The second metric is the time required to compile the random problems, shown in Figure 8.2. Often, Drake's compilation times are much better, but they are highly variable. For a number of the largest problems, Drake is up to about 1000 times faster. However, some problems exhibit little improvement, and a few are up to 100 times slower for the worst cases. Most problems take less than about ten minutes for both methods, but Drake takes a few hours on five of the largest problems. We expect higher variability for the run-time because the compilation algorithms loop repeatedly over the labeled graph, exacerbating the variability shown in the compiled size. Thus, it is not surprising that on problems whose dispatchable form is not compact, the run-time suffers.

The final metric is the run-time latency incurred by the algorithms, shown in Figure 8.3. Drake's reported latency is the maximum latency for a single decision making period during a single execution of the entire problem. The STN enumeration latency reported is the time required to identify, execute, and propagate the first event in each consistent, component STN. While these metrics are not identical, they are quite comparable. The values reported at $10^{-3}$ seconds were reported as zero by Lisp's timing features, and we inflate them to fit on a log scale.

Although differences in compilation time are interesting, increases in the run-time latency are far more critical to Drake's applicability in the real world. Fortunately, the situation looks favorable. Both systems execute all but the largest problems in under a tenth of a second, and most small and moderate sized problems in around 10 milliseconds. Although Drake is slower on a reasonable fraction of the problems, the margin is fairly low; note that the cluster of values at a ratio of just less than $10^{-1}$ corresponds to the jump from effectively zero to about 10 milliseconds. We do not know the real ratio, but these points create a visible cluster at $10^{-1}$ which may or may not be misleading. Instead, we focus on larger problems where both methods are measurable, and where Drake generally performs quite well. Again, a handful of problems are outliers, taking much more time, up to tens of seconds, which would not be acceptable for most applications. We conclude that Drake will perform well on embedded systems for many real world problems, in terms of memory usage and latency.

Overall, these results are what Drake was designed to achieve. Using a compact representation provides a smaller memory footprint. Sometimes, exploiting the similarity between choices makes reasoning fast, and other times it imposes an extra computational burden to tease out the similarities, or lack thereof. Enumerating the STNs directly has quite predictable costs, both in time and space, and Drake is far more variable, depending on the precise nature of the problem. While we find these results quite promising, we must caution that applications to particular problems could be better or worse, depending on factors we do not know how to easily characterize. This is well illustrated by the few problems that were outliers in all three metrics. It may be useful in future work to further investigate the sources of variability in Drake's performance.

As a practical note, we tightly regulated the number of events to focus our investigation on the scaling with the number of choices. However, both algorithms should scale gracefully, and with similar increases in space and time costs, to plans with more events. Overall, Drake appears to provide a noticeably lower memory footprint for dispatching problems with discrete choices than the direct enumeration strategy of Tsamardinos et al. (2001), while only suffering from a mild increase in run-time latency.

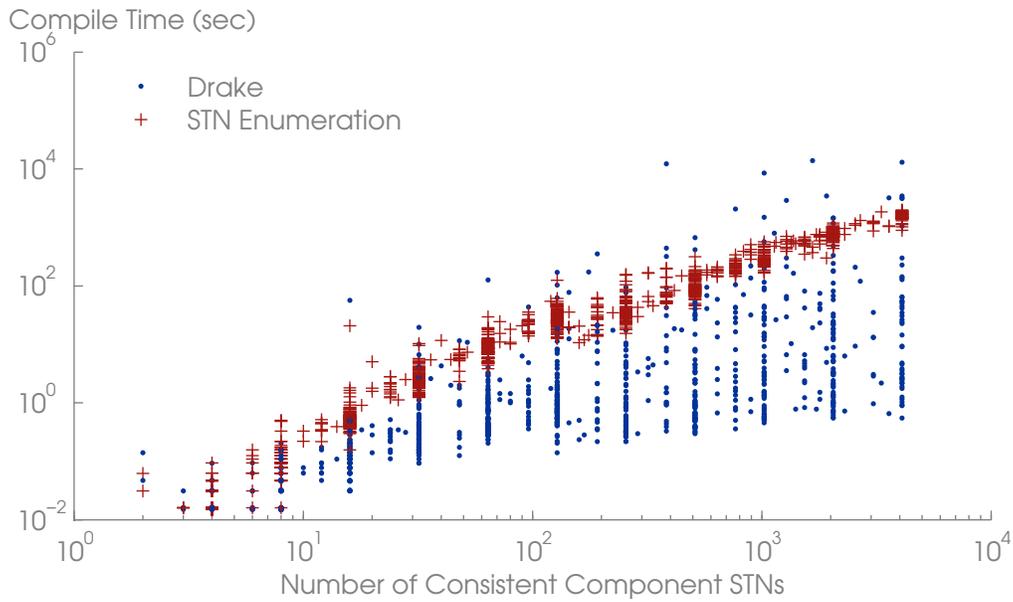

(a) The compile time for random problems.

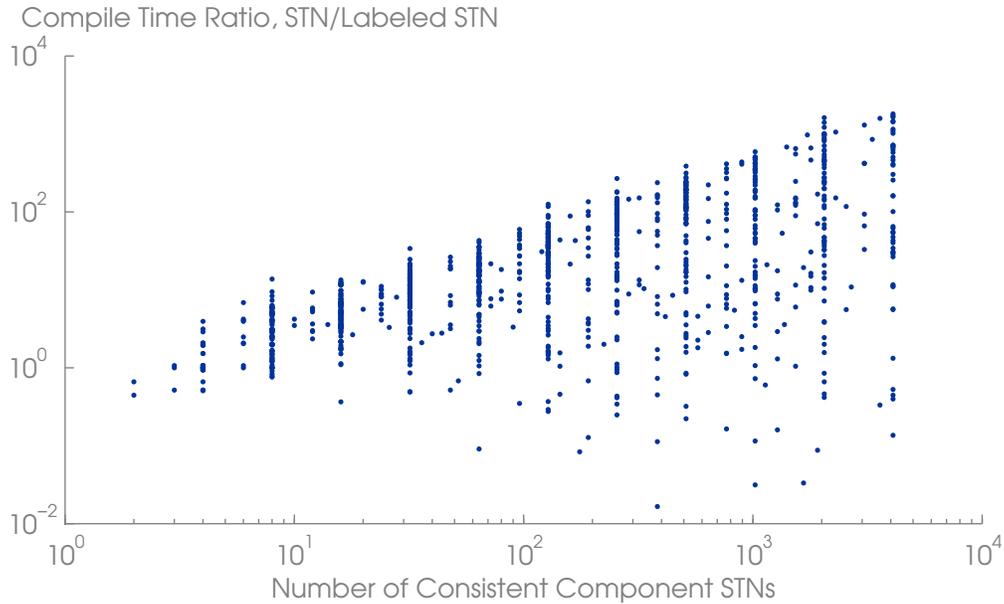

(b) The ratio of the compile time of STN enumeration to Drake's compile time.

Figure 8.2: The compile time of random problems as a function of the number of component STNs.

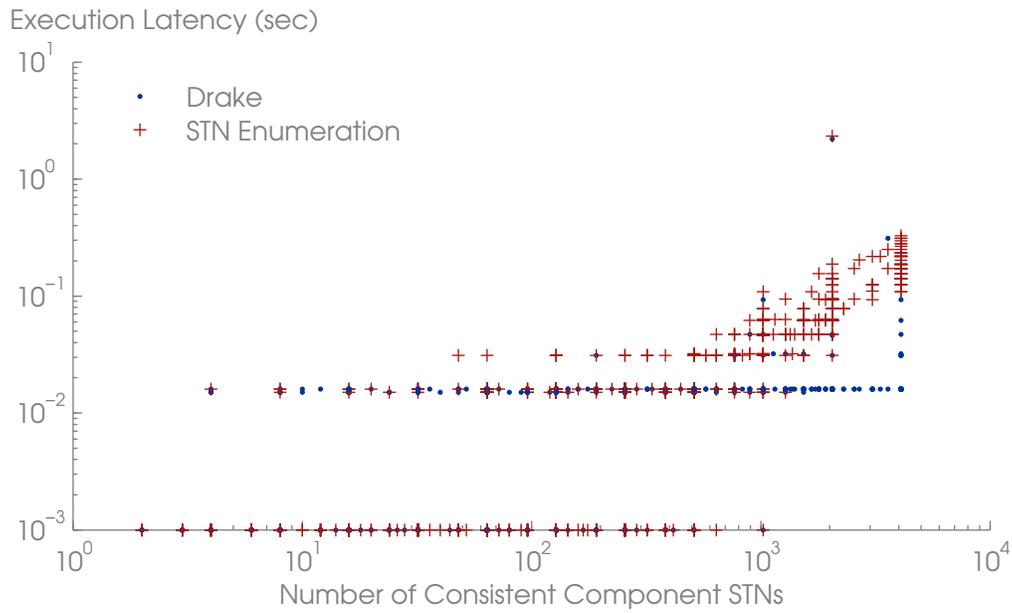

(a) The execution latency.

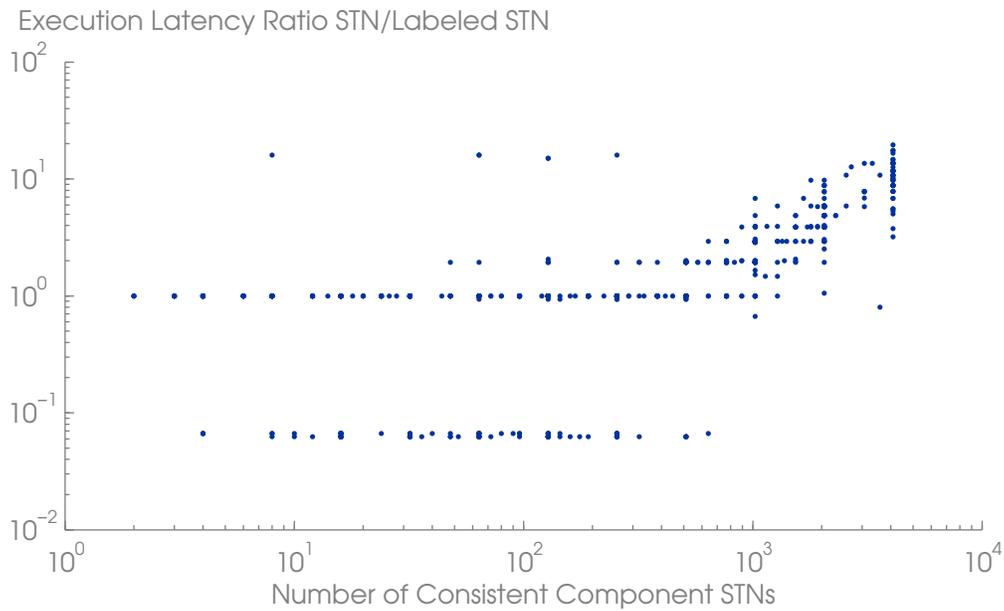

(b) The ratio of the execution latency for STN enumeration to Drake's execution latency.

Figure 8.3: The execution latency of random problems as a function of the number of component STNs.

## 9. Summary

This work presents Drake, a compact, flexible executive for plans with choice. Drake takes input plans with temporal flexibility and discrete choices, such as Labeled STNs or DTNs, and selects the execution times and makes discrete decisions at run-time (Dechter et al., 1991). Choices substantially improve the expressiveness of the tasks that executives can perform, and improve the robustness of the resulting executions. Prior execution approaches typically impose significant memory requirements or introduce substantial latency during execution. Our goal in developing Drake is to develop a dispatching executive with a lower memory footprint.

Building upon the concept of labels employed by the ATMS to compactly encode all the consequences of a set of alternative choices, Drake introduces a new compact encoding, called labeled distance graphs, to encode and efficiently reason over discrete choices, and we introduce a corresponding maintenance system (de Kleer, 1986). Our adaptation of the ATMS labeling scheme focuses on only maintaining non-dominated constraints, which allows Drake to exploit the structure of temporal reasoning, cast as a shortest path problem on a distance graph, to provide a compact representation. Furthermore, modifying the existing unlabeled algorithms to account for labels does not change the overall structure of the algorithms.

Drake's compilation algorithm successfully compresses the dispatchable solution by over two orders of magnitude relative to Tsamardinos, Pollack, and Ganchev's (2001) prior work, often reducing the compilation time, and typically introducing only a modest increase in execution latency. Thus, we believe that Drake successfully realizes our initial goals. Within our experiments, compilation typically takes less than ten minutes, but on occasion takes hours. Although time consuming in the later case, this is still acceptable, since compilation can be performed off-line, when a task is first defined. To summarize, Drake's Labeled STNs and labeled distance graphs enable an executive that strikes a useful balance between latency and memory consumed, which is appropriate for real world applications. Drake's labeling scheme also provides the opportunity to extend a wide range of graph algorithms to reason about and represent choice efficiently.

## Acknowledgments

The authors would like to thank Julie Shah and David Wang for many helpful ideas and discussions, and the reviewers for their insightful comments. Patrick Conrad was funded during this work by a Department of Defense NDSEG Fellowship.